\documentclass[10pt]{IEEEtran}

\usepackage{cite}
\usepackage{graphicx}
\usepackage{subfigure}
\usepackage{amsmath}  
\usepackage{amsfonts}
\usepackage[latin1]{inputenc}
\usepackage{pstricks}

\DeclareFontFamily{OT1}{mvs}{}
\DeclareFontShape{OT1}{mvs}{m}{n}{<-> fmvr8x}{}
\def\mvs{\usefont{OT1}{mvs}{m}{n}}
\def\mvchr{\mvs\char}
\def\Squarepipe{{\mvchr151}}

\begin{document}
%
\title{Retinal Vessel Segmentation Using the 2-D Morlet Wavelet and
  Supervised Classification}
%
%
\author{João~V.~B.~Soares,
        Jorge~J.~G.~Leandro,
        Roberto~M.~Cesar-Jr.,
        Herbert~F.~Jelinek,
        and Michael~J.~Cree,~\IEEEmembership{Senior~Member,~IEEE}
\thanks{©2006 IEEE. Personal use of this material is permitted. However, permission to reprint/republish this material for advertising or promotional purposes or for creating new collective works for resale or redistribution to servers or lists, or to reuse any copyrighted component of this work in other works must be obtained from the IEEE.}
\thanks{This work was supported by CNPq (131403/2004-4, 300722/98-2 and 474596/2004-4), FAPESP (99/12765-2), the Australian Diabetes Association and the CSU CoS.}
\thanks{J. Soares, J. Leandro, and R. Cesar-Jr. are with the Institute of Mathematics and Statistics - University of São Paulo - USP, Brazil (e-mails: \{joao, jleandro, cesar\}@vision.ime.usp.br).}
\thanks{H. Jelinek is with the School of Community Health, Charles Sturt University, Australia (e-mail: hjelinek@csu.edu.au).}
\thanks{M. Cree is with the Department of Physics and Electronic Engineering, University of Waikato, Hamilton, New Zealand (e-mail: cree@waikato.ac.nz).}}

\maketitle

\begin{abstract}
  We present a method for automated segmentation of the vasculature in
  retinal images. The method produces segmentations by classifying
  each image pixel as {\it vessel} or {\it non-vessel}, based on the
  pixel's feature vector. Feature vectors are composed of the pixel's
  intensity and continuous two-dimensional Morlet wavelet transform
  responses taken at multiple scales. The Morlet wavelet is capable of
  tuning to specific frequencies, thus allowing noise filtering and
  vessel enhancement in a single step. We use a Bayesian classifier
  with class-conditional probability density functions (likelihoods)
  described as Gaussian mixtures, yielding a fast classification,
  while being able to model complex decision surfaces and compare its
  performance with the linear minimum squared error classifier.  The
  probability distributions are estimated based on a training set of
  labeled pixels obtained from manual segmentations.  The method's
  performance is evaluated on publicly available DRIVE~\cite{staal04}
  and STARE~\cite{hoover00} databases of manually labeled
  non-mydriatic images. On the DRIVE database, it achieves an area
  under the receiver operating characteristic (ROC) curve of 0.9598,
  being slightly superior than that presented by the method of Staal
  {\it et al.}~\cite{staal04}.
\end{abstract}


\IEEEpeerreviewmaketitle

\section{Introduction}

\begin{figure}
  \centerline{
    \subfigure[Inverted green channel of non-mydriatic fundus image.]{
      \includegraphics[width=.23\textwidth]{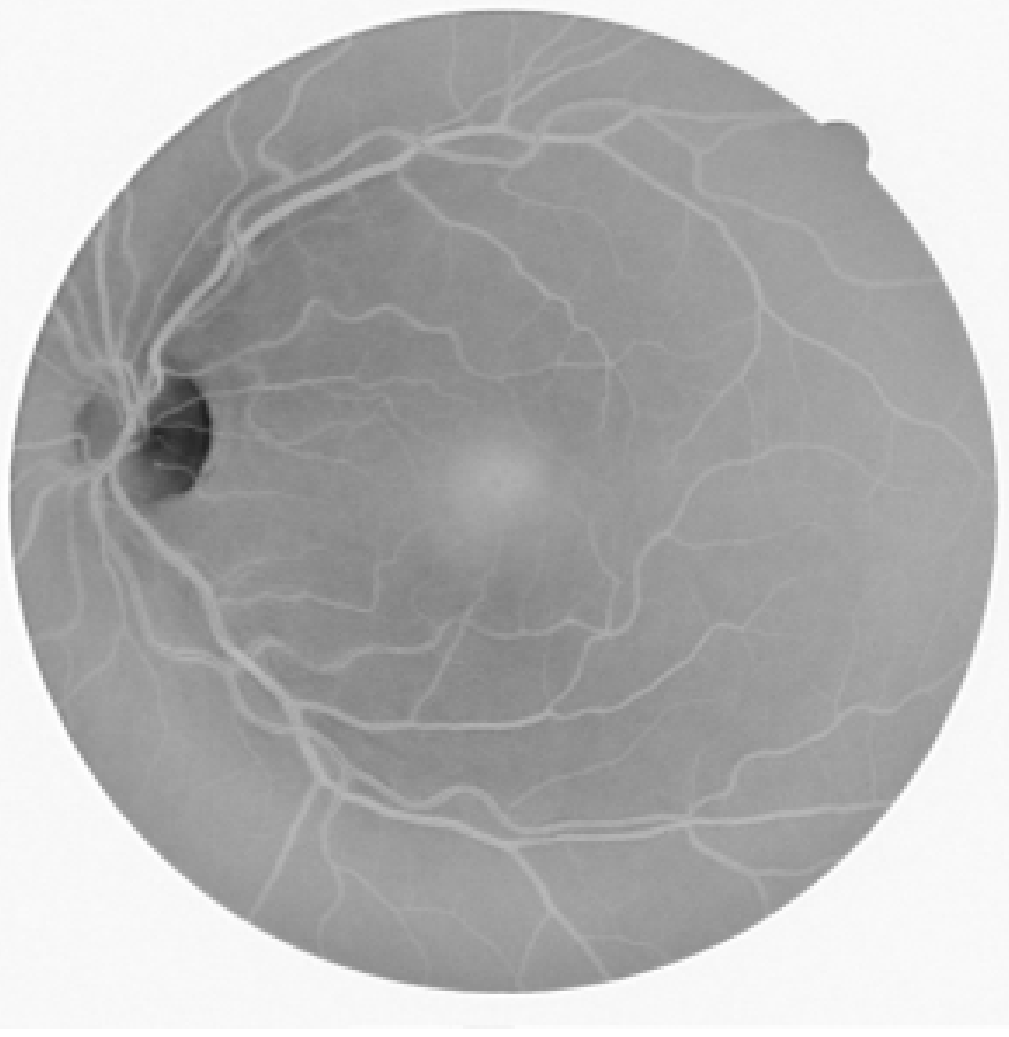}
      \label{original}}
    \hfil
    \subfigure[Pre-processed image with extended border. The original image limit is indicated for illustration.]{
      \includegraphics[width=.23\textwidth]{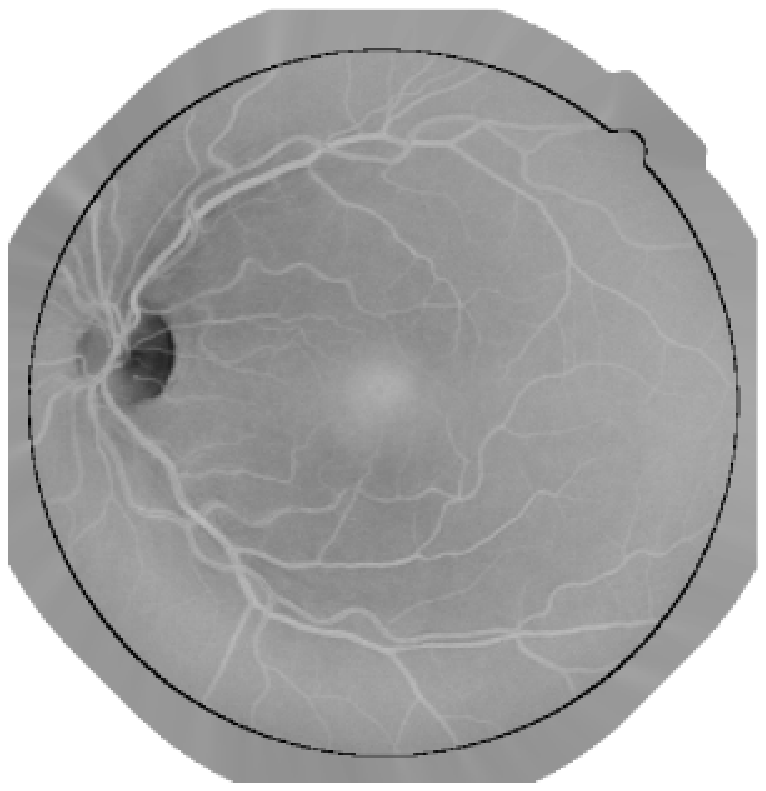}
      \label{padded}
    }
  }
  \caption{Fundus image pre-processing for removing undesired border effects.}
  \label{pre}
\end{figure}

\PARstart{O}{ptic} fundus (Fig.~\ref{original}) assessment has been
widely used by the medical community for diagnosing vascular and
non-vascular pathology. Inspection of the retinal vasculature may
reveal hypertension, diabetes, arteriosclerosis, cardiovascular
disease and stroke~\cite{kanski89}. Diabetic retinopathy is a major
cause of adult blindness due to changes in blood vessel structure and
distribution such as new vessel growth (proliferative diabetic
retinopathy) and requires laborious analysis from a
specialist~\cite{sussman82}. Endeavoring to reduce the effect of
proliferative diabetic retinopathy includes obtaining and analyzing
images of the optic fundus at regular intervals such as every six
months to a year. Early recognition of changes to the blood vessel
patterns can prevent major vision loss as early intervention becomes
possible~\cite{lee01, taylor01}.

To provide the opportunity for initial assessment to be carried out by
community health workers, computer based analysis has been introduced,
which includes assessment of the presence of microaneurysms and
changes in the blood flow/vessel distribution due to either vessel
narrowing, complete occlusions or new vessel
growth~\cite{mcquellin02, streeter-c-2003, wong05}.

An automatic assessment for blood vessel anomalies of the optic fundus
initially requires the segmentation of the vessels from the
background, so that suitable feature extraction and processing may be
performed. Several methods have been developed for vessel
segmentation, but visual inspection and evaluation by receiver
operating characteristic (ROC) analysis shows that there is still room
for improvement \cite{niemeijer04, cree05}. In addition, it is
important to have segmentation algorithms that do not critically
depend on configuring several parameters so that untrained community
health workers may utilize this technology. These limitations of the
state-of-the-art algorithms have motivated the development of the
framework described here, which only depends on manually segmented
images.

Many different approaches for automated vessel segmentation have been
reported. The papers \cite{liu93, zhou94, chutatape98, toliaspanas98,
  can99, lalonde00, gao01} present vessel tracking methods to obtain
the vasculature structure, along with vessel diameters and branching
points. Tracking consists of following vessel center lines guided by
local information, usually trying to find the path which best matches
a vessel profile model. The use of deformable models also shows
promising results in~\cite{mcinerney00, toledo00, vasilevskiy02,
  nain04}. In~\cite{chaudhuri89, hoover00, gang02}, matched filters
are used to emphasize blood vessels. An improvement is obtained
in~\cite{hoover00} by a region-based threshold probing of the matched
filter response. Multithreshold probing is directly applied to the
images in~\cite{jiangmojon03}. A non-linear filter that enhances
vessels by exploiting properties of the vessel profiles is introduced
in~\cite{lowell04}. Along this line is the use of mathematical
morphology filtering in~\cite{zana01, fang03}, coupled with curvature
evaluation. In~\cite{martinezperez00}, multi-scale curvature and
border detection are used to drive a region growing algorithm.

Supervised methods for pixel classification have been shown
in~\cite{nekovei95, sinthanayothin99, staal04}. In~\cite{nekovei95},
feature vectors are formed by gray-scale values from a window centered
on the pixel being classified. A window of values is also used
in~\cite{sinthanayothin99}, but the features used are a principal
component transformation of RGB values and edge strength.
In~\cite{staal04}, ridge detection is used to form line elements and
partition the image into patches belonging to each line element. Pixel
features are then generated based on this representation. Many
features are presented and a feature selection scheme is used to
select those which provide the best class separability.

Previously, we have shown promising preliminary results using the
continuous wavelet transform (CWT)~\cite{leandro01, cesarjelinek03}
and integration of multi-scale information through supervised
classification~\cite{leandro03}. Here we improve on those methods
using a Bayesian classifier with Gaussian mixture models as class
likelihoods and evaluate performances with ROC analysis. ROC analysis
has been used for evaluation of segmentation methods
in~\cite{hoover00, jiangmojon03, staal04} and comparison of some of
the cited methods in~\cite{niemeijer04, cree05}.

In our approach, each pixel is represented by a feature vector
including measurements at different scales taken from the continuous
two-dimensional Morlet wavelet transform. The resulting feature space
is used to classify each pixel as either a {\it vessel} or {\it
  non-vessel} pixel. We use a Bayesian classifier with
class-conditional probability density functions (likelihoods)
described as Gaussian mixtures, yielding a fast classification, while
being able to model complex decision surfaces and compare its
performance with the linear minimum squared error classifier.

Originally devised for suitably analyzing non-stationary and
inhomogeneous signals, the time-scale analysis took place to
accomplish unsolvable problems within the Fourier framework, based on
the continuous wavelet transform (CWT).  The CWT is a powerful and
versatile tool that has been applied to many different image
processing problems, from image coding~\cite{rioul91} to shape
analysis~\cite{costabook01}. This success is largely due to the fact
that wavelets are especially suitable for detecting singularities
(e.g. edges and other visual features) in images~\cite{grossmann88},
extracting instantaneous frequencies~\cite{antoine93}, and performing
fractal and multi-fractal analysis.  Furthermore, the wavelet transform
using the Morlet wavelet, also often referred to as Gabor wavelet, has
played a central role in increasing our understanding of visual
processing in different contexts from feature detection to face
tracking~\cite{feris04}. The Morlet wavelet is directional and capable
of tuning to specific frequencies, allowing it to be adjusted for
vessel enhancement and noise filtering in a single step. These nice
characteristics motivate the adoption of the Morlet wavelet in our
proposed framework.

This work is organized as follows. The databases used for tests are
described in Subsection~\ref{materials}. Subsection~\ref{methods}
presents our segmentation framework based on supervised pixel
classification. In Subsection~\ref{features} the feature generation
process is described, including the 2-D CWT and Morlet wavelet. Our
use of supervised classification and the classifiers tested are
presented in Subsection~\ref{supervised}. ROC analysis for performance
evaluation is described in Subsection~\ref{evaluation} and results are
presented in Section~\ref{results}. Discussion and conclusion are in
Section~\ref{discussion}.

\section{Materials and methods}

\subsection{Materials}
\label{materials}

There are different ways of obtaining ocular fundus images, such as
with non-mydriatic cameras, which do not require the dilation of the
eyes through drops, or through angiograms using fluorescein as a
tracer~\cite{lee01}. We have tested our methods on angiogram
gray-level images and colored non-mydriatic images~\cite{leandro01,
  leandro03}. Here, our methods are tested and evaluated on two
publicly available databases of non-mydriatic images and corresponding
manual segmentations: the DRIVE~\cite{staal04} and
STARE~\cite{hoover00} databases.

The DRIVE database consists of 40 images (7 of which present
pathology), along with manual segmentations of the vessels. The images
are captured in digital form from a Canon CR5 non-mydriatic 3CCD camera
at $45^\circ$ field of view (FOV). The images are of size $768 \times
584$ pixels, $8$ bits per color channel and have a FOV of
approximately $540$ pixels in diameter. The images are in compressed
JPEG-format, which is unfortunate for image processing but is commonly
used in screening practice.

The 40 images have been divided into a training and test set, each
containing 20 images (the training set has 3 images with pathology).
They have been manually segmented by three observers trained by an
ophthalmologist. The images in the training set were segmented once,
while images in the test set were segmented twice, resulting in sets A
and B. The observers of sets A and B produced similar segmentations.
In set A, 12.7\% of pixels where marked as vessel, against 12.3\%
vessel for set B. Performance is measured on the test set using the
segmentations of set A as ground truth. The segmentations of set B are
tested against those of A, serving as a human observer reference for
performance comparison.

The STARE database consists of 20 digitized slides captured by a
TopCon TRV-50 fundus camera at $35^\circ$ FOV. The slides were
digitized to $700 \times 605$ pixels, $8$ bits per color channel.  The
FOV in the images are approximately $650 \times 550$ pixels in
diameter. Ten of the images contain pathology. Two observers manually
segmented all images. The first observer segmented 10.4\% of pixels as
vessel, against 14.9\% vessels for the second observer. The
segmentations of the two observers are fairly different in that the
second observer segmented much more of the thinner vessels than the
first one. Performance is computed with the segmentations of the first
observer as ground truth.

\subsection{General framework}

\label{methods}

The image pixels of a fundus image are viewed as objects represented
by feature vectors, so that we may apply statistical classifiers in
order to segment the image. In this case, two classes are considered,
i.e. {\it vessel} $\times$ {\it non-vessel} pixels.  The training set
for the classifier is derived by manual segmentations of training
images, i.e. pixels segmented by hand are labeled as {\it vessel}
while the remaining pixels are labeled as {\it non-vessel}.  This
approach allows us to integrate information from wavelet responses at
multiple scales in order to distinguish pixels from each class.

\subsection{Pixel features}

\label{features}

When the RGB components of the non-mydriatic images are visualized
separately, the green channel shows the best vessel/background
contrast (Fig.~\ref{original}), whereas, the red and blue channels
show low contrast and are very noisy. Therefore, the green channel was
selected to be processed by the wavelet, as well as to compose the
feature vector itself, i.e. the green channel intensity of each pixel
is taken as one of its features. For angiograms, the wavelet is
applied directly to the gray-level values, which are also used to
compose the feature vectors.

\subsubsection{Pre-processing}

In order to reduce false detection of the border of the camera's
aperture by the wavelet transform, an iterative algorithm has been
developed. Our intent is to remove the strong contrast between the
retinal fundus and the region outside the aperture (see
Fig.~\ref{pre}).

The pre-processing algorithm consists of determining the pixels
outside the aperture that are neighbors to pixels inside the aperture
and replacing each of their values with the mean value of their
neighbors inside the aperture. This process is repeated and can be
seen as artificially increasing the area inside the aperture, as shown
in Fig.~\ref{padded}.

Before the application of the wavelet transform to non-mydriatic
images, we invert the green channel of the image, so that the vessels
appear brighter than the background.

\subsubsection{Wavelet transform features}

The notation and definitions in this section follow~\cite{arneodo00}.
The real plane $\mathbb{R} \times \mathbb{R}$ is denoted as
$\mathbb{R}^2$, and the vectors are represented as bold letters, e.g.
$\mathbf{x}, \mathbf{b} \in \mathbb{R}^2$. Let $f\in L^2$ be an image
represented as a square integrable (i.e. finite energy) function
defined over $\mathbb{R}^2$.  The continuous wavelet transform $T_\psi
(\mathbf{b},\theta, a)$ is defined as:

\begin{equation}
\label{eq-wvlt}
\nonumber
T_\psi (\mathbf{b},\theta ,a) = C_\psi ^{-1/2} \frac{1}{a}\int {\psi ^\ast ({a^{-1}r_{-\theta }( {\mathbf{x}-\mathbf{b}})} )f(\mathbf{x})d^2\mathbf{x}} 
\end{equation}

\noindent where $ C_\psi$, $\psi$, $\mathbf{b}$, $\theta$ and $a$
denote the normalizing constant, analyzing wavelet, the displacement
vector, the rotation angle and the dilation parameter (also known as
scale), respectively.  $\psi^\ast$ denotes the complex conjugate of
$\psi$.

Combining the conditions for both the analyzing wavelet and its
Fourier transform of being well localized in the time and frequency
domain plus the requirement of having zero mean, one realizes that the
wavelet transform provides a local filtering at a constant rate
$\frac{\Delta \omega}{\omega} $, indicating its great efficiency as
the frequency increases, i.e. as the scale decreases.  This property
is what makes the wavelet effective for detection and analysis of
localized properties and singularities~\cite{antoine93}, such as the
blood vessels in the present case.

Among several available analyzing wavelets, for instance, the 2-D
Mexican hat and the optical wavelet, we chose the 2-D Morlet wavelet
for the purposes of this work, due to its directional selectiveness
capability of detecting oriented features and fine tuning to specific
frequencies~\cite{antoine93, arneodo00}. This latter property is
especially important in filtering out the background noise of the
fundus images. The 2-D Morlet wavelet is defined as:

\begin{equation}
\label{morlet}
\nonumber
\psi_M (\mathbf{x}) = \exp(j \mathbf{k_0} \mathbf{x}) \exp\left( - \frac{1}{2}\vert A \mathbf{x} \vert^2 \right)
\end{equation}

\noindent
where $j=\sqrt{ -1 }$ and $A = \mbox{diag} [\epsilon^{- 1/2} , 1],
\epsilon \geq 1$ is a $2 \times 2$ diagonal matrix that defines the
anisotropy of the filter, i.e. its elongation in any desired
direction~\cite{antoine93}. The Morlet wavelet is actually a complex
exponential modulated Gaussian, where $\mathbf{k}_0$ is a vector that
defines the frequency of the complex exponential.

\begin{figure}
  \hspace{10pt}
  \subfigure[Surface representation of the real part.]{
    \includegraphics[width=.42\textwidth]{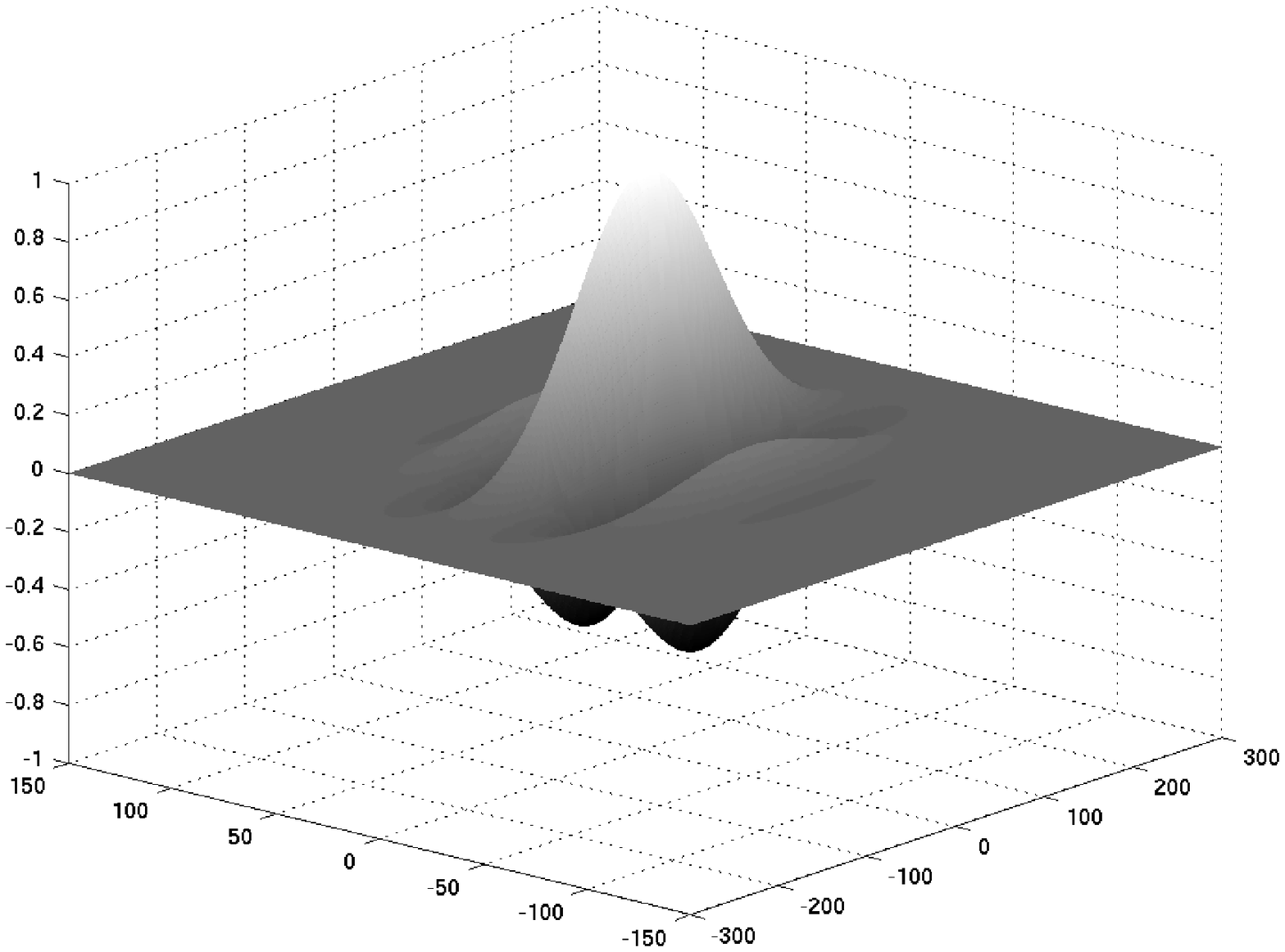}
    \put(-180,5){\footnotesize $x$}
    \put(-40,5){\footnotesize $y$}
    \rput{90}(-7.8,3){\footnotesize $\psi_M(x,y)$}
    \label{wvlt_3d}
  }
  \centerline{
    \subfigure[Real part.]{
      \includegraphics[width=.22\textwidth]{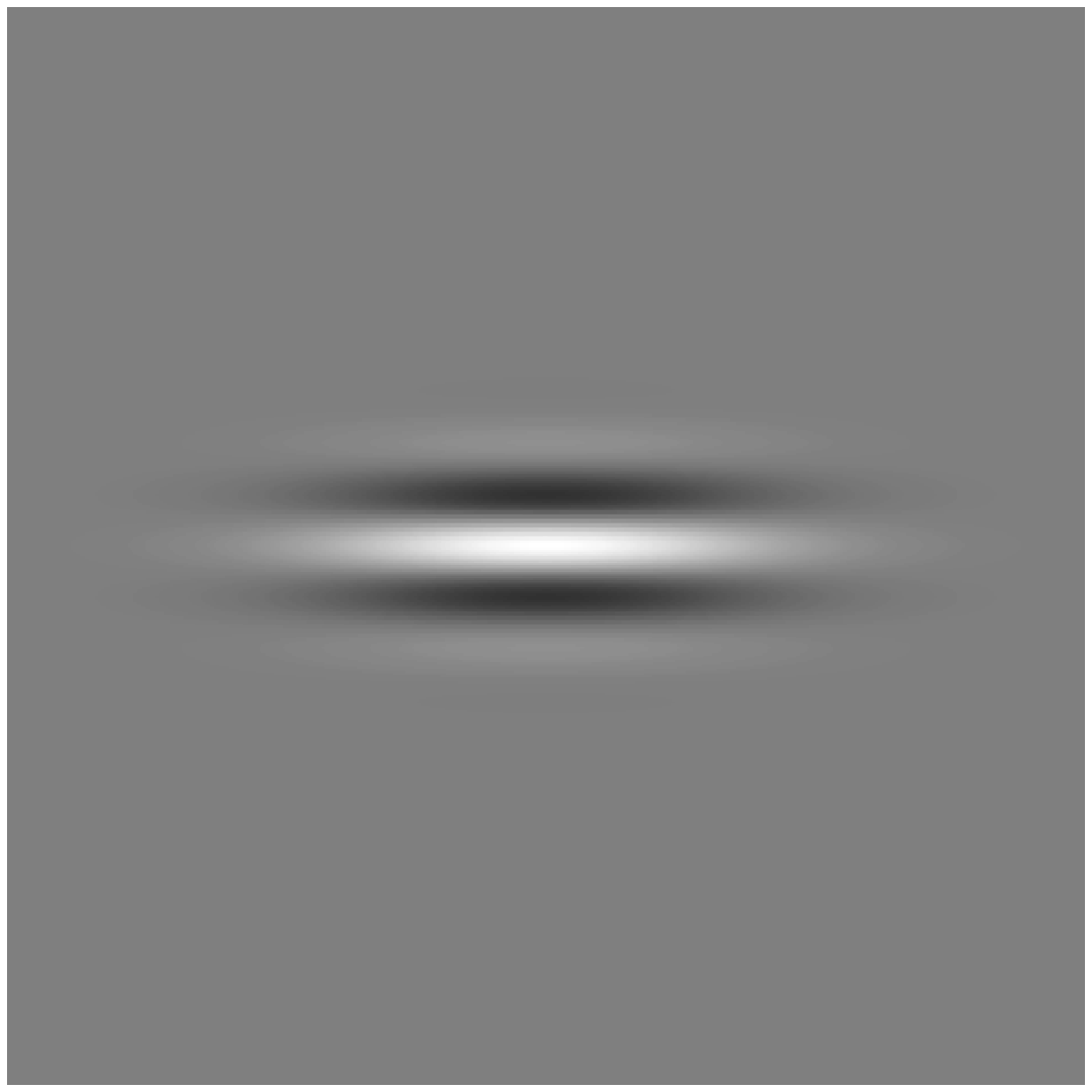}
      \label{wvlt_real}
    }
    \hfill
    \subfigure[Imaginary part.]{
      \includegraphics[width=.22\textwidth]{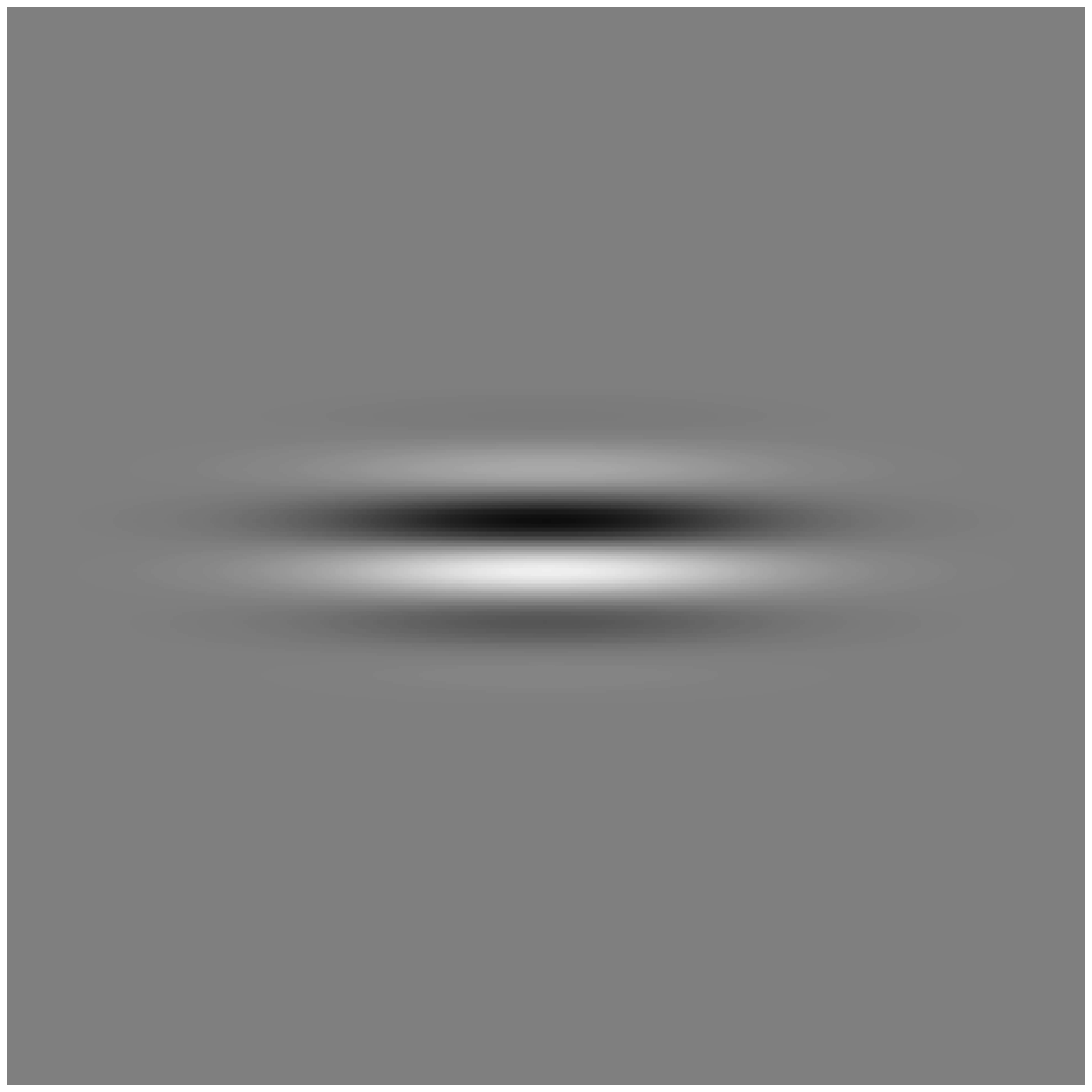}
      \label{wvlt_imag}
    }
  }
  \caption{Different representations for the 2-D Morlet wavelet ($\psi_M$) with parameters $\epsilon = 8$ and $\mathbf{k}_0 = [0, 3]$.}
  \label{wavelet}
\end{figure}

We have set the $\epsilon$ parameter to $8$, making the filter
elongated and $\mathbf{k_0} = [0 , 3]$, i.e. a low frequency complex
exponential with few significant oscillations, as shown in
Fig.~\ref{wavelet}.  These two characteristics have been chosen in
order to enable the transform to present stronger responses for pixels
associated with the blood vessels.

For each considered scale value, we are interested in the response
with maximum modulus over all possible orientations, i.e.:

\begin{equation}
\label{eq-wvlt-max}
M_\psi (\mathbf{b},a) = \max_{\theta}|T_\psi (\mathbf{b},\theta ,a)|
\end{equation}

\begin{figure}[t]
  \centerline{
    \subfigure[$M_\psi (\mathbf{b},2)$.]{\includegraphics[width=.23\textwidth]{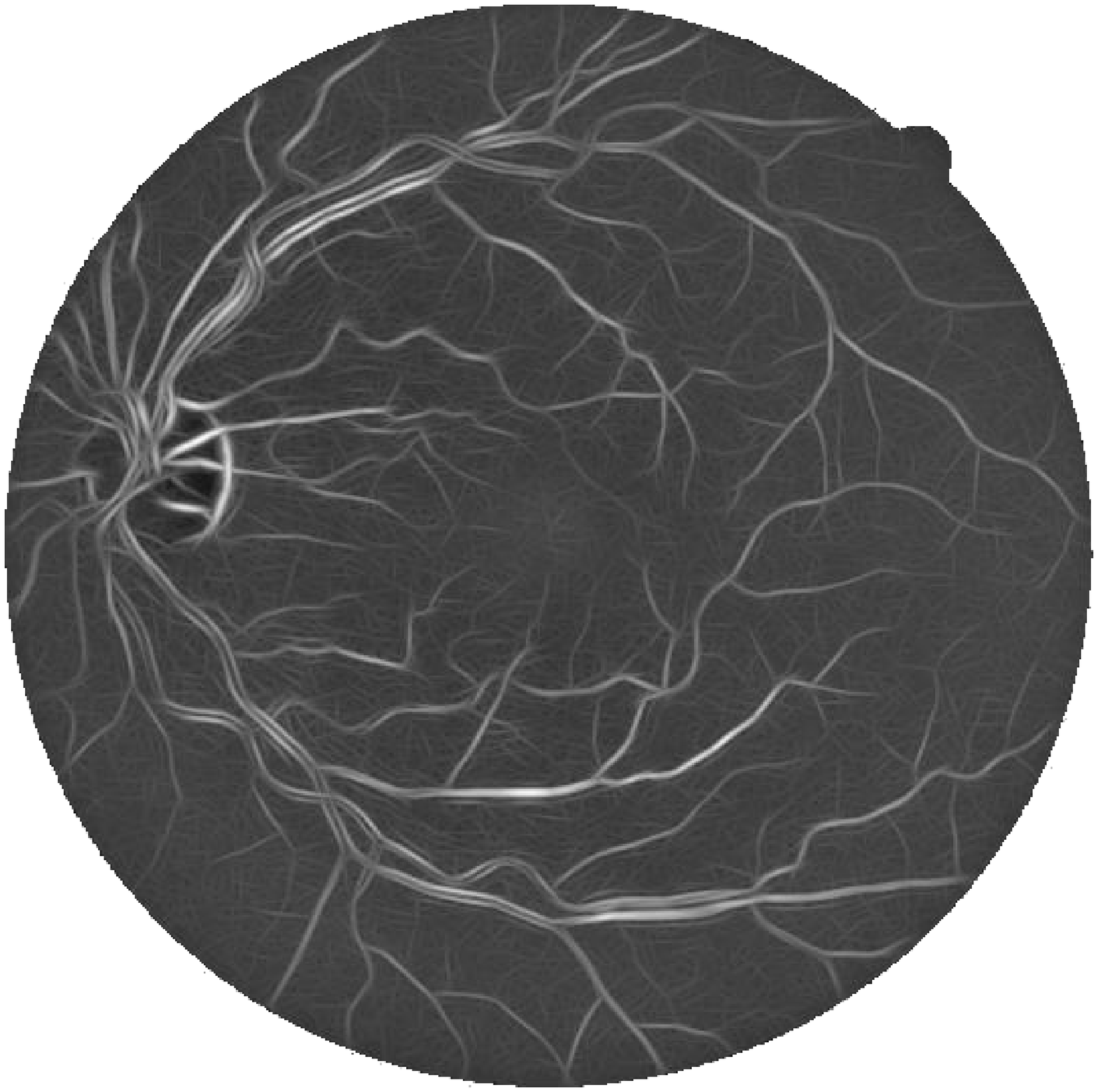}
      \label{first}
    }
    \hfil
    \subfigure[$M_\psi (\mathbf{b},4)$.]{\includegraphics[width=.23\textwidth]{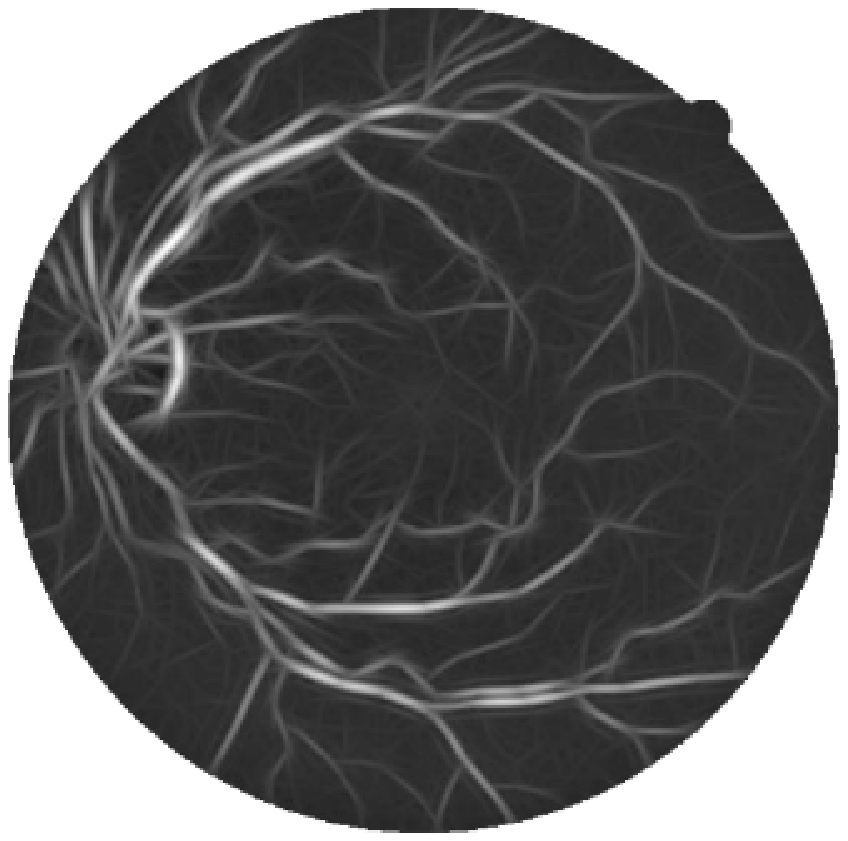}
      \label{fig_second}
    }
  }
  \caption{Maximum modulus of Morlet wavelet transform over angles,
    $M_\psi (\mathbf{b},a)$ (Eq.~\ref{eq-wvlt-max}), for scale values
    of $a = 2$ and $a = 4$ pixels. The remaining parameters are fixed
    at $\epsilon = 8$ and $\mathbf{k}_0 = [0, 3]$.}
  \label{transform}
\end{figure}

Thus, the Morlet wavelet transform is computed for $\theta $ spanning
from $0$ up to $170$ degrees at steps of $10$ degrees and the maximum
is taken (this is possible because $|T_\psi (\mathbf{b},\theta ,a)| =
|T_\psi (\mathbf{b},\theta + 180,a)|$). The maximum modulus of the
wavelet transform over all angles for multiple scales are then taken
as pixel features. $M_\psi (\mathbf{b},a)$ is shown in
Fig.~\ref{transform} for $a = 2$ and $a = 4$ pixels.

\subsubsection{Feature normalization}

Given the dimensional nature of the features forming the feature
space, one must bear in mind that this might give rise to errors in
the classification process, as the units chosen might affect the
distance in the feature space.

A strategy to obtain a new random variable with zero mean and unit
standard deviations, yielding, in addition, dimensionless features, is
to apply the normal transformation to the feature space.  The normal
transformation is defined as~\cite{costabook01}:

\begin{equation}
\nonumber
\hat{v}_i = \frac{v_i - \mu_i}{\sigma_i}
\end{equation}

\noindent where $v_i$ is the $i^{th}$ feature assumed by each pixel,
$\mu_i$ is the average value of the $i^{th}$ feature and $\sigma_i$ is
the associated standard deviation.

We have applied the normal transformation separately to each image's
feature space, i.e., every image's feature space is normalized by its
own means and standard deviations, helping to compensate for intrinsic
variation between images (e.g. illumination).

\subsection{Supervised classification for segmentation}

\label{supervised}

Supervised classification has been applied to obtain the final
segmentation, with the pixel classes defined as
$C_1=\lbrace\emph{vessel pixels}\rbrace$ and
$C_2=\lbrace\emph{non-vessel pixels}\rbrace$. In order to obtain the
training set, several fundus image have been manually segmented,
allowing the creation of a labeled training set into classes $C_1$ and
$ C_2$ (see Subsection~\ref{materials}). Due to the computational cost
of training the classifiers and the large number of samples, we
randomly select a subset of the available samples to use for actually
training the classifiers. We will present results for two different
classifiers, described below.

\subsubsection{Gaussian mixture model Bayesian classifier}
We have achieved very good results using a Bayesian classifier in
which each class-conditional probability density function (likelihood)
is described as a linear combination of Gaussian functions~\cite{
  theodoridis99, dhs01}. We will call this the {\it Gaussian mixture
  model} (GMM) classifier.

The {\it Bayes classification rule} for a feature vector $\mathbf{v}$
can be stated in terms of posterior probabilities as

\begin{equation}
  \label{decision}
  \begin{array}{l} 
    \mbox{Decide } C_1 \mbox{ if } P(C_1 | \mathbf{v}) > P(C_2 | \mathbf{v});\\
    \mbox{otherwise, decide } C_2
  \end{array}
\end{equation}

We recall {\it Bayes rule}:
\begin{equation}
\label{eq-bayes}
  P(C_i | \mathbf{v}) = \frac{p(\mathbf{v}|C_i) P(C_i)}{p(\mathbf{v})}
\end{equation}

\noindent where $p(\mathbf{v}|C_i)$ is the class-conditional
probability density function, also known as likelihood, $P(C_i)$ is
the prior probability of class $C_i$, and $p(\mathbf{v})$ is the
probability density function of $\mathbf{v}$ (sometimes called
evidence).

To obtain a decision rule based on estimates from our training set, we
apply {\it Bayes rule} to Eq.~\ref{decision}, obtaining the equivalent
decision rule:

\begin{equation}
  \nonumber  
  \begin{array}{l}
    \mbox{Decide } C_1 \mbox{ if } p(\mathbf{v} | C_1) P(C_1) > p(\mathbf{v} | C_2) p(C_2);\\ 
    \mbox{otherwise, decide } C_2
  \end{array}
\end{equation}

We estimate $P(C_i)$ as $N_i / N$, the ratio of class $i$ samples in
the training set. The class likelihoods are described as linear
combinations of Gaussian functions:

\begin{equation}
\nonumber
  p(\mathbf{v} | C_i) = \sum_{j=1}^{k_i} p(\mathbf{v} | j, C_i) P_j
\end{equation}

\noindent where $k_i$ is the number of Gaussians modeling likelihood
$i$, $P_j$ is the weight of Gaussian $j$ and each $p(\mathbf{v} | j,
C_i)$ is a $d$-dimensional Gaussian distribution.

For each class $i$, we estimate the $k_i$ Gaussian parameters and
weights with the Expectation-Maximization (EM)
algorithm~\cite{theodoridis99}. The EM algorithm is an iterative
scheme that guarantees a local maximum of the likelihood of the
training data.

GMMs represent a halfway between purely nonparametric and parametric
models, providing a relatively fast classification process at the cost
of a more expensive training algorithm.

\subsubsection{Linear minimum squared error classifier}

We have also tested the linear minimum squared error
classifier~\cite{dhs01, theodoridis99}, denoted LMSE. Linear
classifiers are defined by a linear decision function $g$ in the
$d$-dimensional feature space:

\begin{equation}
\label{eq-lin}
  g(\mathbf{v}) = \mathbf{w}^t \mathbf{v} + w_0
\end{equation}

\noindent where $\mathbf{v}$ is a feature vector, $\mathbf{w}$ is the
weight vector and $w_0$ the threshold.

The classification rule is to decide $C_1$ if $g(\mathbf{v}) > 0$ and
$C_2$ otherwise. To simplify the formulation, the threshold $w_0$ is
accommodated by defining the extended $(d + 1)$-dimensional vectors
$\mathbf{v}' \equiv [\mathbf{v}^T, 1]^T$ and $\mathbf{w}' \equiv
[\mathbf{w}^T, w_0]^T$, so that $g(\mathbf{v}) = \mathbf{w}'^T
\mathbf{v}'$.

The classifier is determined by finding $\mathbf{w}'$ that minimizes
the {\it sum of error squares} criterion:

\begin{equation}
\nonumber
  J(\mathbf{w}') = \sum_{i=1}^{N}(y_i - \mathbf{v}_i'^T \mathbf{w}')^2
\end{equation}

\noindent where $N$ is the total number of training samples,
$\mathbf{v}'_i$ is the extended $i^{th}$ training sample, and $y_i$
its desired output.

The criterion measures the sum of squared errors between the true
output of the classifier ($\mathbf{v}_i'^T \mathbf{w}'$) and the
desired output ($y_i$). We have arbitrarily set $y_i = 1$ for
$\mathbf{v}_i \in C_1$ and $y_i = - 1$ for $\mathbf{v}_i \in C_2$.

Let us define

\begin{equation}
\nonumber
V = 
\left[
\begin{array}{c} 
\mathbf{v}_1'^T\\ 
\mathbf{v}_2'^T\\
\vdots\\
\mathbf{v}_N'^T
\end{array} 
\right]
,\hspace{15pt}
\mathbf{y} = 
\left[
\begin{array}{c} 
y_1\\ 
y_2\\
\vdots\\
y_N
\end{array} 
\right]
\end{equation}

Minimizing the criterion with respect to $\mathbf{w}'$ results in:

\begin{equation}
\nonumber
  (V^T V)\hat{\mathbf{w}}' = V^T \mathbf{y} \Rightarrow \hat{\mathbf{w}}' = (V^T V)^{-1}V^T \mathbf{y}
\end{equation}

In comparison to the GMM classifier, the LMSE classifier has a much
faster training process, but is restricted in the sense that it is
linear, while GMMs allow for complex decision boundaries. However, as
we will show, the results obtained using LMSE are comparable to those
using GMMs, representing a reasonable trade-off.

\subsection{Experimental evaluation}
\label{evaluation}

The performances are measured using receiver operating characteristic
(ROC) curves. ROC curves are plots of true positive fractions versus
false positive fractions for varying thresholds on the posterior
probabilities. A pair formed by a true positive fraction and a false
positive fraction is plotted on the graph for each threshold value (as
explained below), producing a curve as in Figs.~\ref{roc}
and~\ref{roc-stare}. The true positive fraction is determined by
dividing the number of true positives by the total number of vessel
pixels in the ground truth segmentations, while the false positive
fraction is the number of false positives divided by the total number
of non-vessel pixels in the ground truth. In our experiments, these
fractions are calculated over all test images, considering only pixels
inside the FOV.

For the GMM classifier, the ROC curve is produced by varying the
threshold on the posterior pixel probabilities (see
Eq.~\ref{eq-bayes}), while the LMSE ROC curve is produced varying the
threshold $w_0$ on the projection of the feature vectors on the
discriminant vector (see Eq.~\ref{eq-lin}).

We have tested our methods on the DRIVE and STARE databases with the
following settings. The pixel features used for classification were
the inverted green channel and its maximum Morlet transform response
over angles $M_{\psi}(\mathbf{b}, a)$ (Eq.~\ref{eq-wvlt-max}) for
scales $a = 2, 3, 4, 6$ pixels (see Subsection~\ref{features}).

For the DRIVE database, the training set was formed by pixel samples
from the 20 labeled training images. For the STARE database,
leave-one-out tests where performed, i.e., every image is segmented
using samples from the other 19 images for the training set. Due to
the large number of pixels, in all experiments, one million pixel
samples where randomly chosen to train the classifiers. Tests were
performed with the LMSE and GMM classifiers. For the GMM classifier,
we vary the number $k = k_1 = k_2$ of {\it vessel} and {\it
  non-vessel} Gaussians modeling each class likelihood.

\section{Results}
\label{results}

\begin{figure*}
  \centerline{\subfigure[Posterior probabilities.]{\includegraphics[width=.23\textwidth]{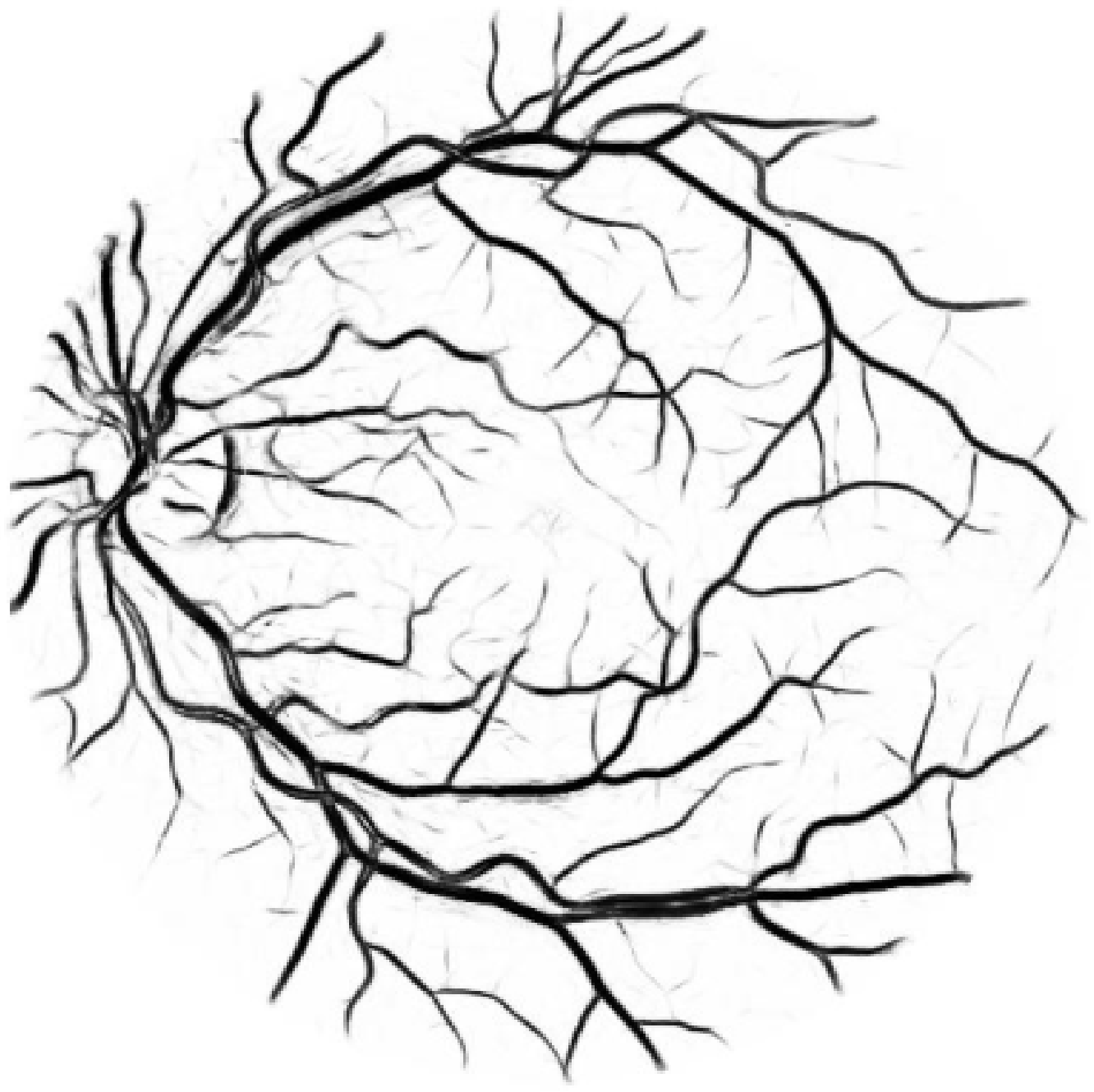}
      \label{posterior}}
    \hfil
    \subfigure[Segmentation.]{\includegraphics[width=.23\textwidth]{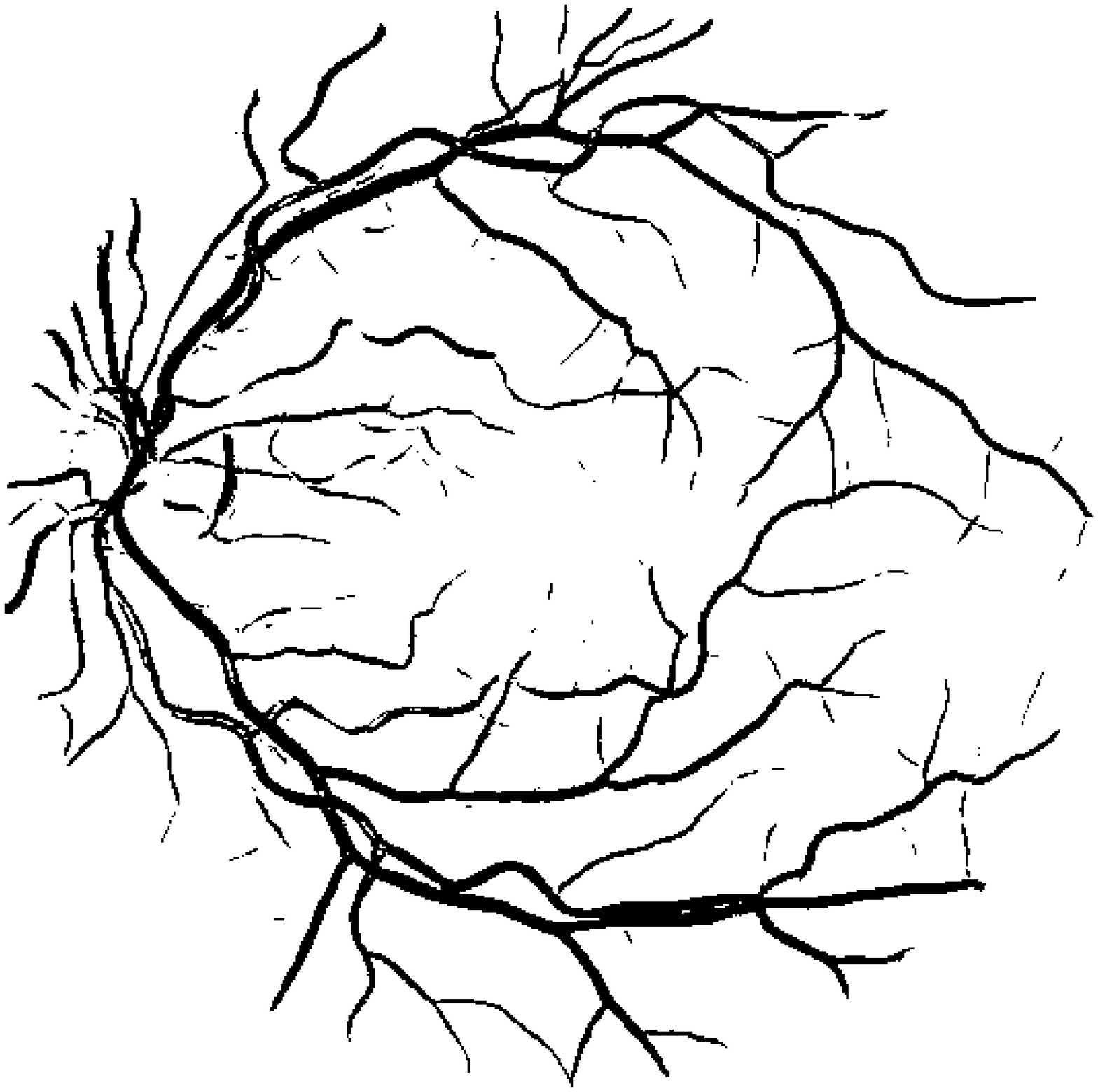}
      \label{segmented}}
    \hfil
    \subfigure[Set A.]{\includegraphics[width=.23\textwidth]{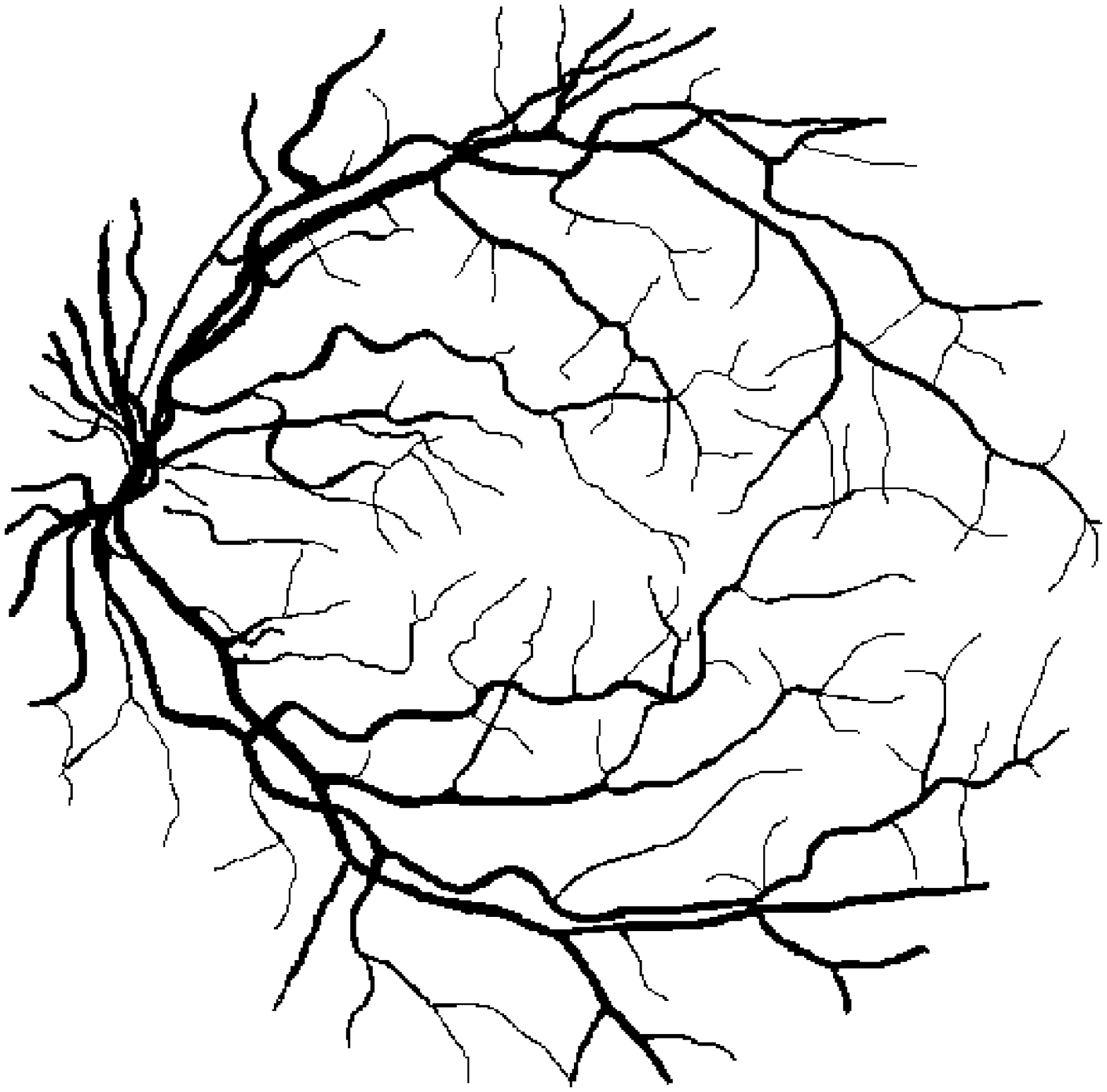}
      \label{manual1}}
    \subfigure[Set B.]{\includegraphics[width=.23\textwidth]{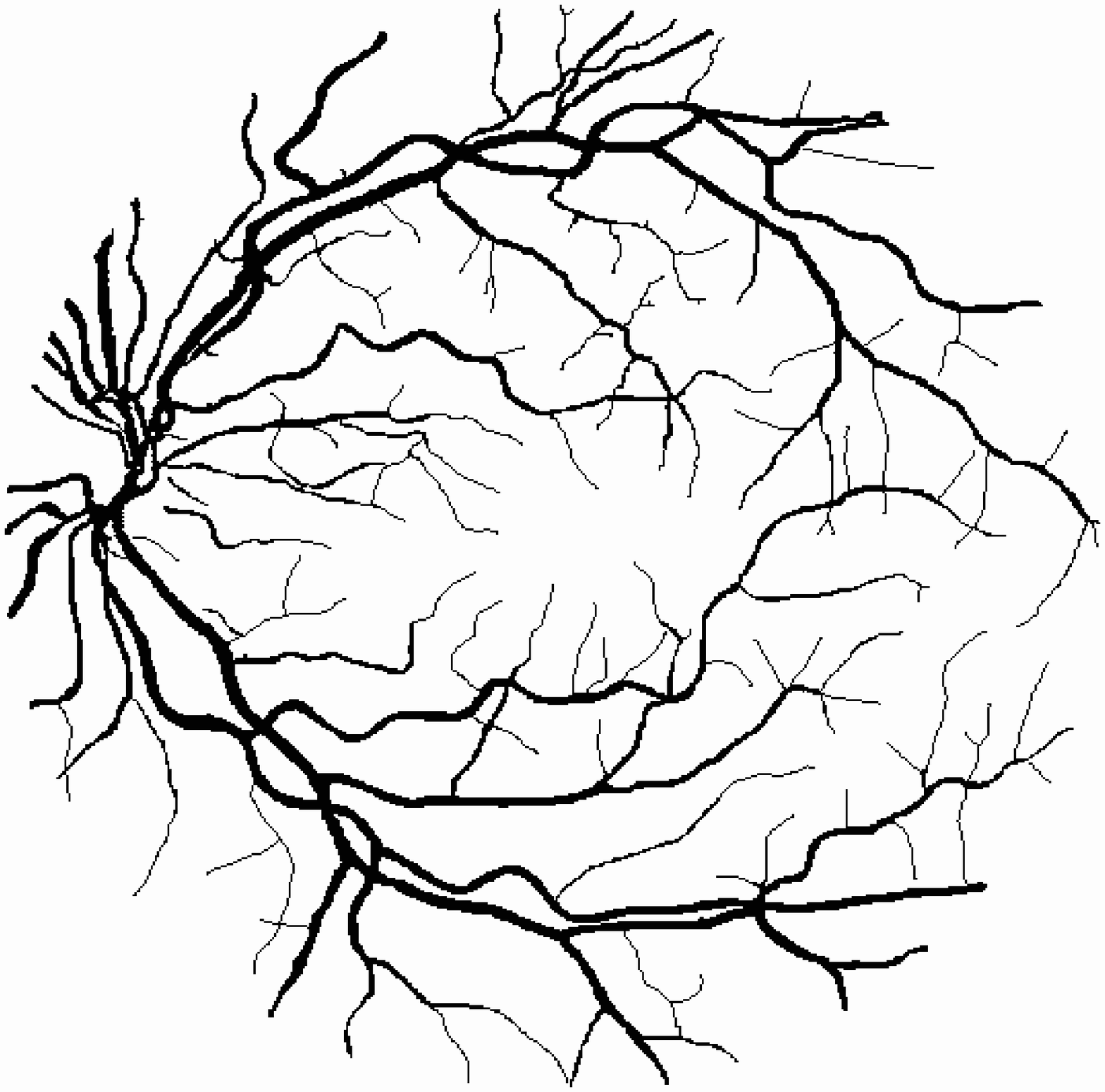}
      \label{manual2}}
  }

  \centerline{\subfigure[Posterior probabilities.]{\includegraphics[width=.23\textwidth]{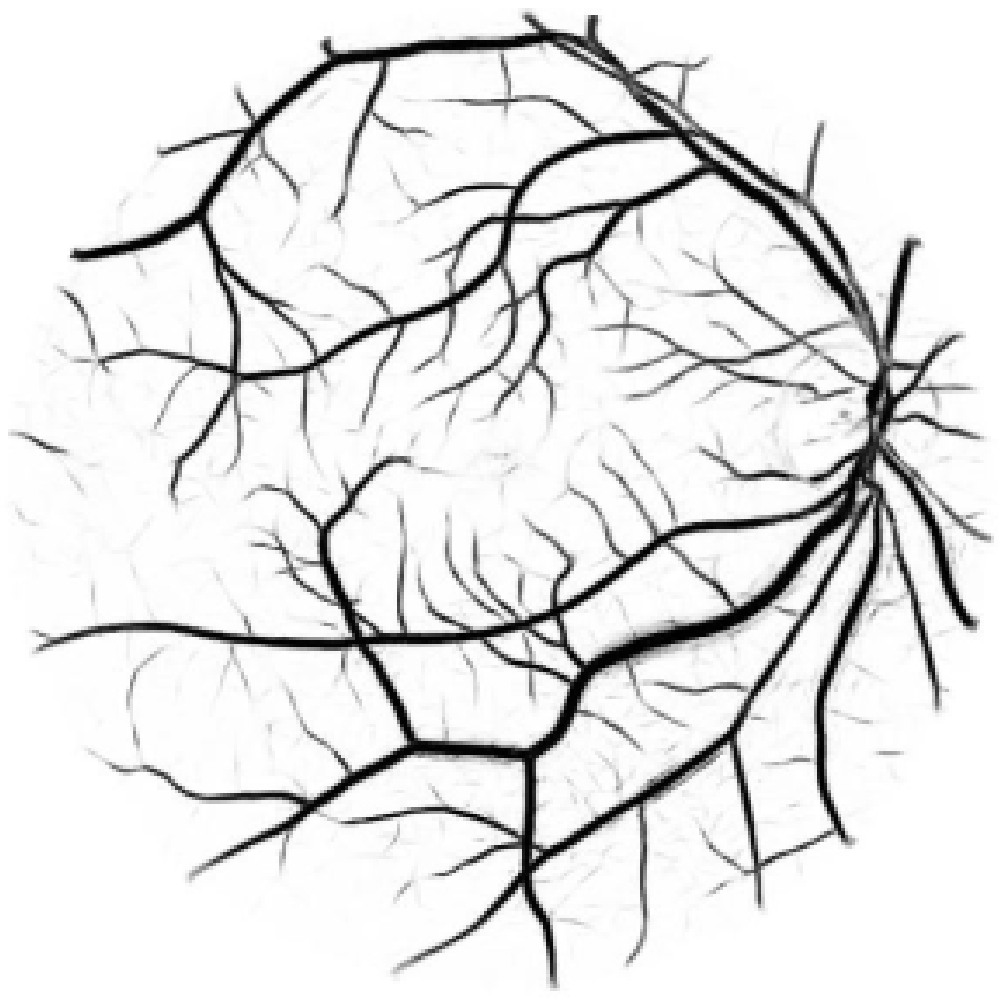}
      \label{posterior16}}
    \hfil
    \subfigure[Segmentation.]{\includegraphics[width=.23\textwidth]{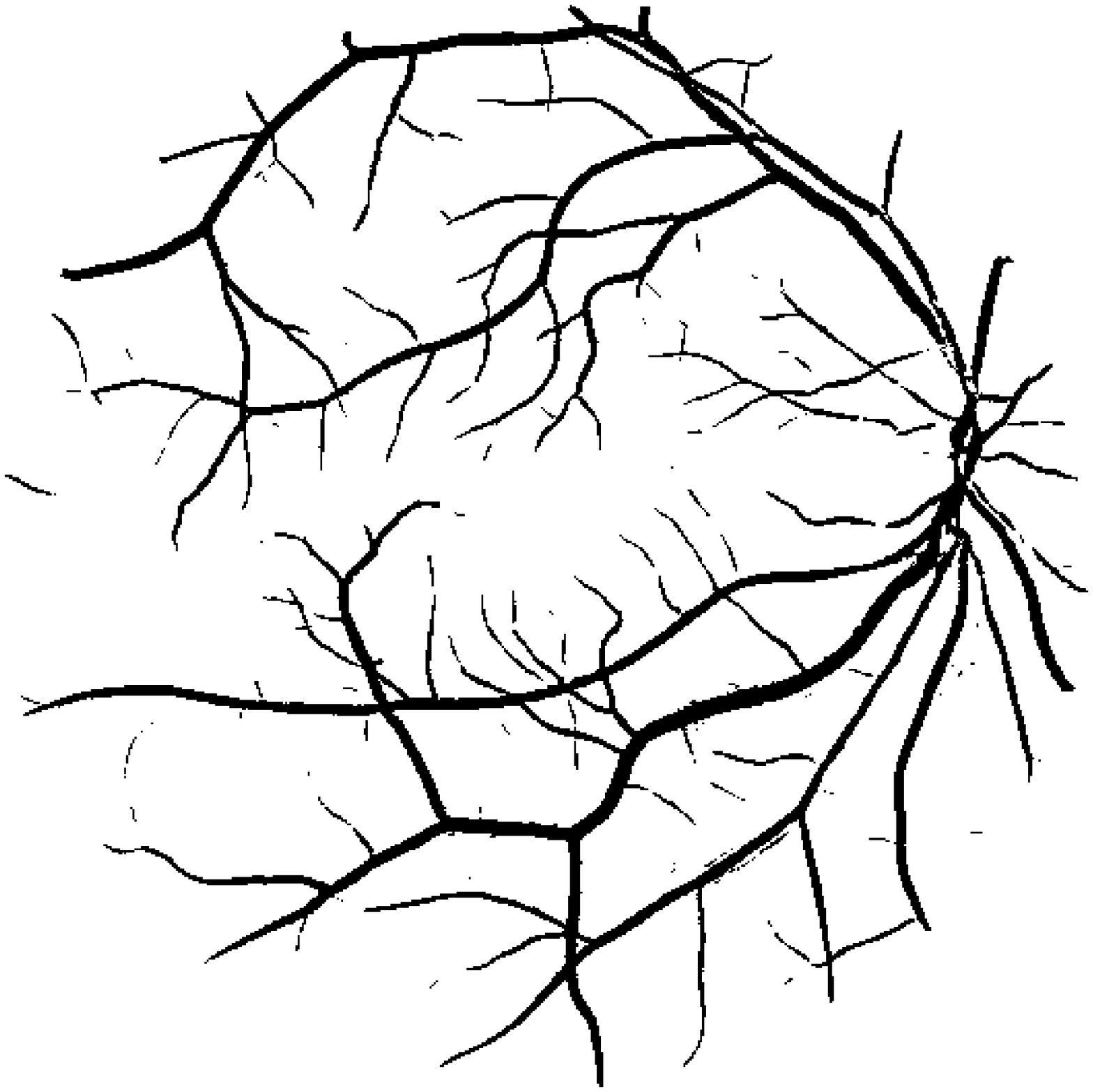}
      \label{segmented16}}
    \hfil
    \subfigure[Set A.]{\includegraphics[width=.23\textwidth]{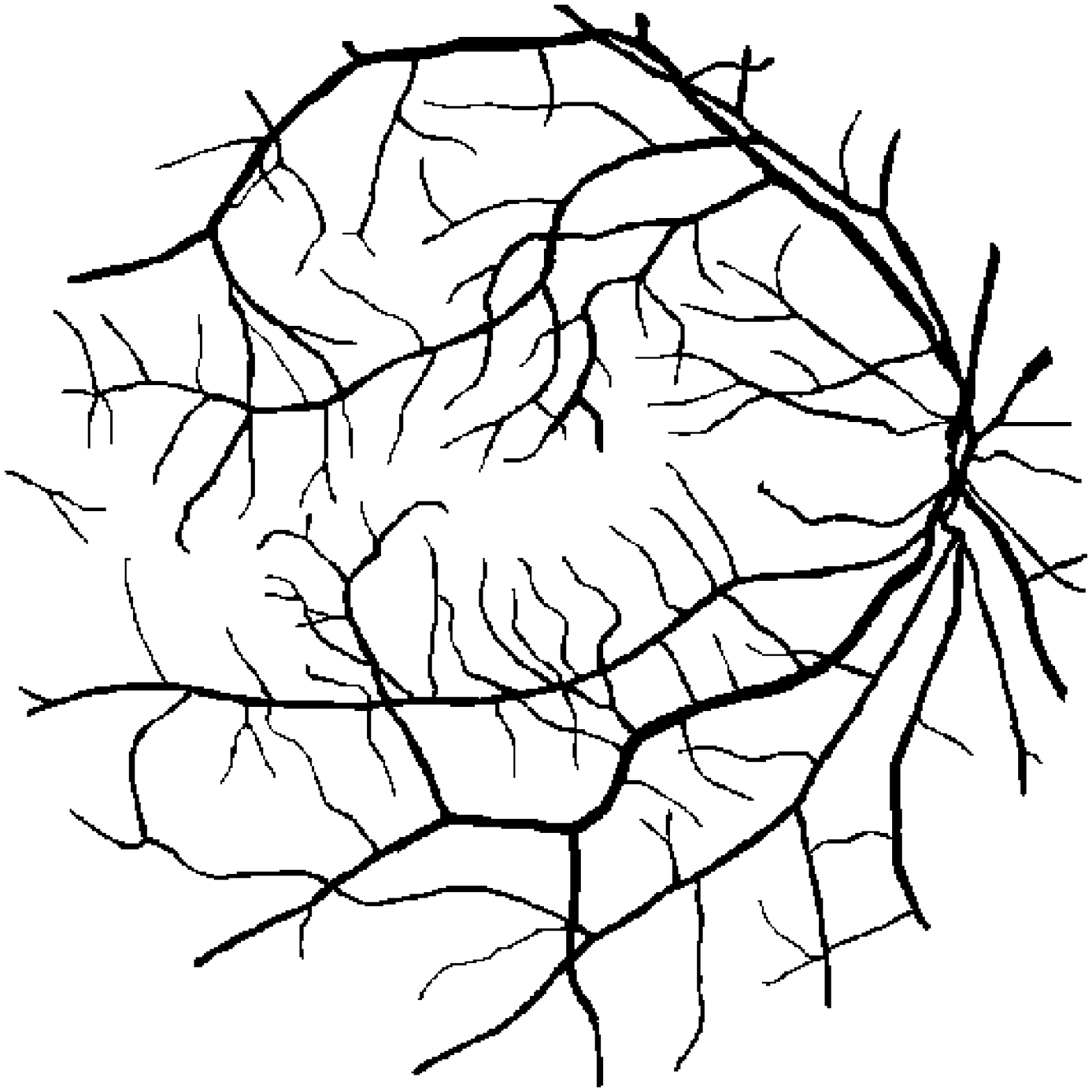}
      \label{manual116}}
    \subfigure[Set B.]{\includegraphics[width=.23\textwidth]{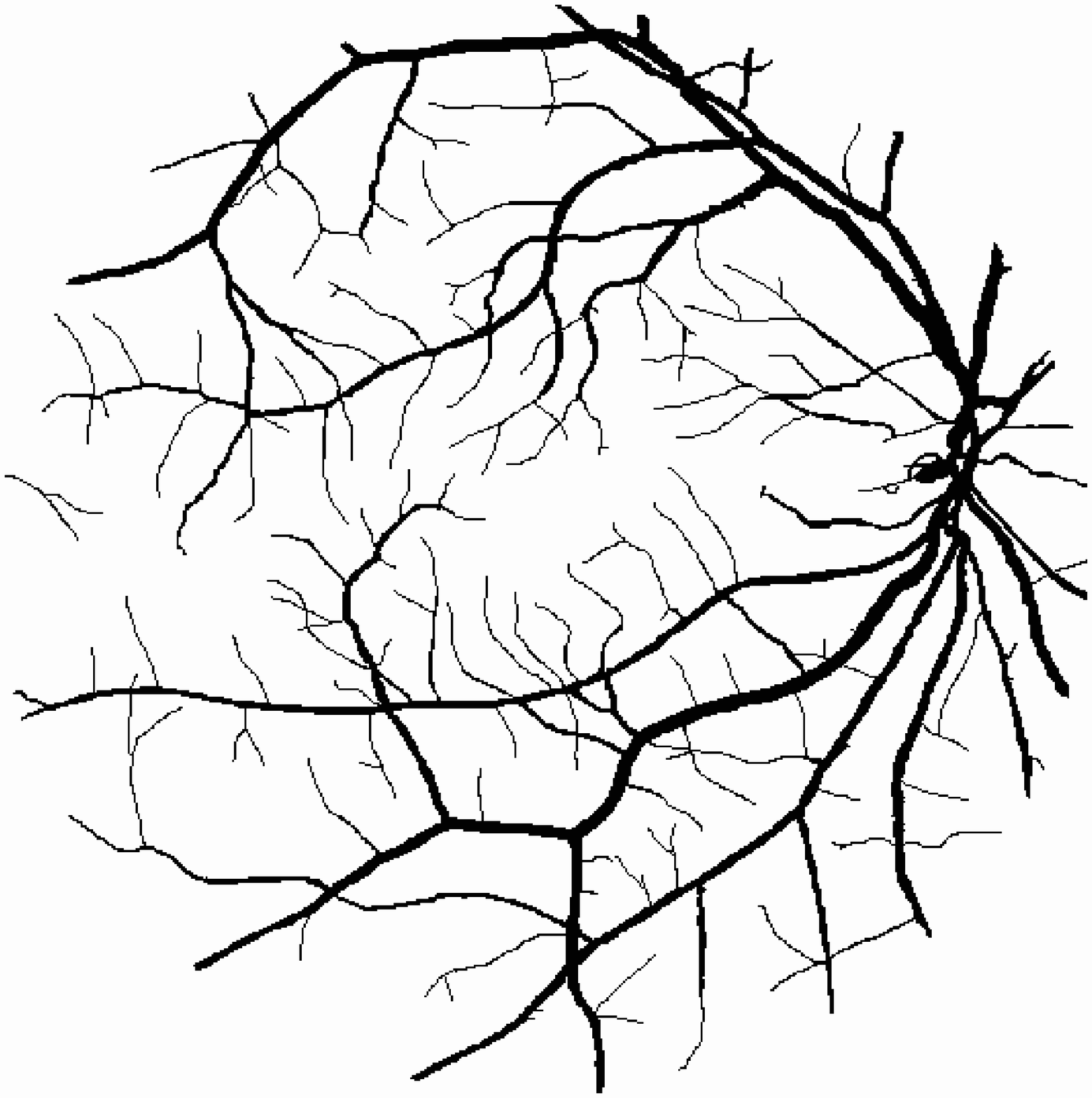}
      \label{manual216}}
  }

  \caption{Results produced by the GMM classifier with $k = 20$ and
    manual segmentations for two images from the DRIVE database. The
    top row results are for the image shown in Fig.~\ref{original}.
  }
  \label{result-drive}
\end{figure*}

\begin{figure*}
  \centerline{\subfigure[Posterior probabilities.]{\includegraphics[width=.23\textwidth]{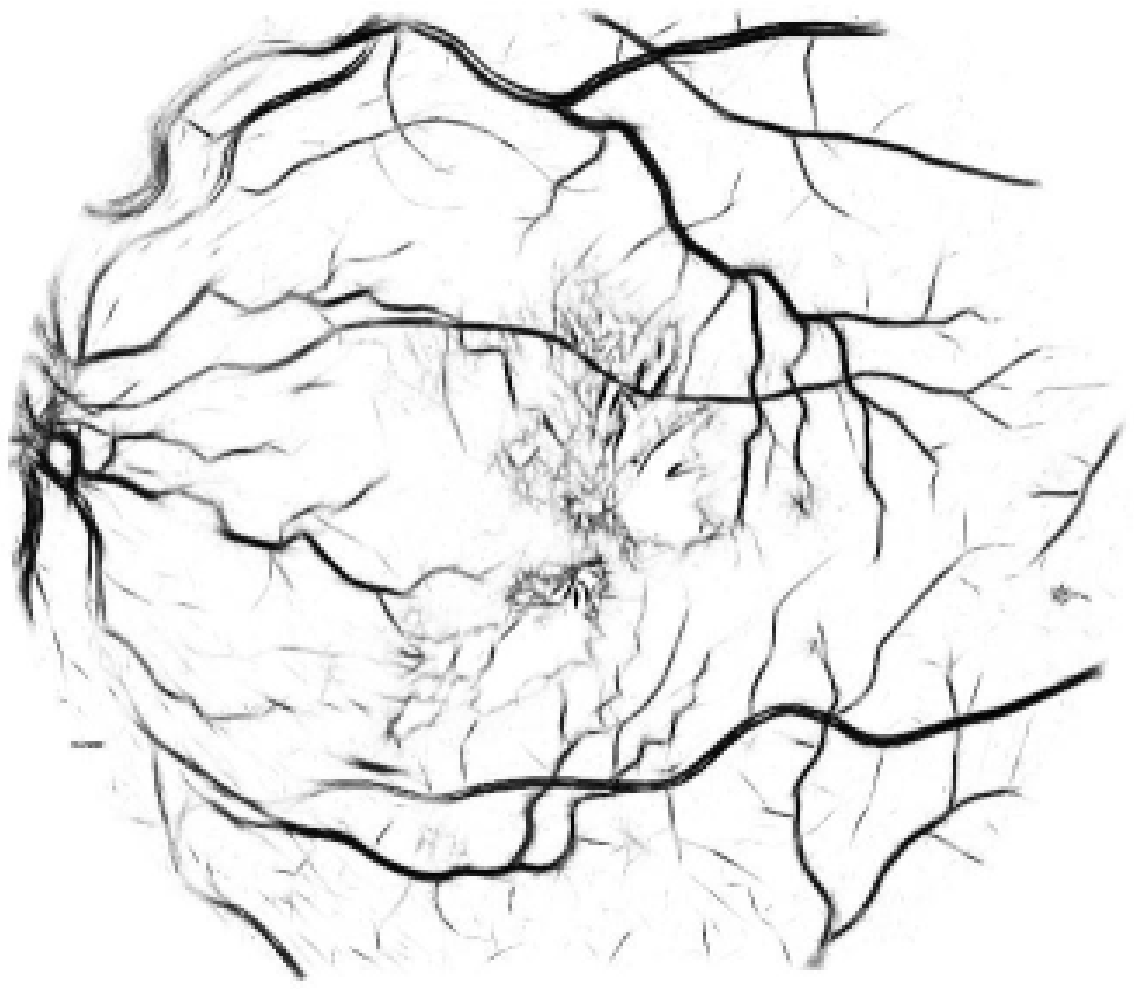}
    }
    \hfil
    \subfigure[Segmentation.]{\includegraphics[width=.23\textwidth]{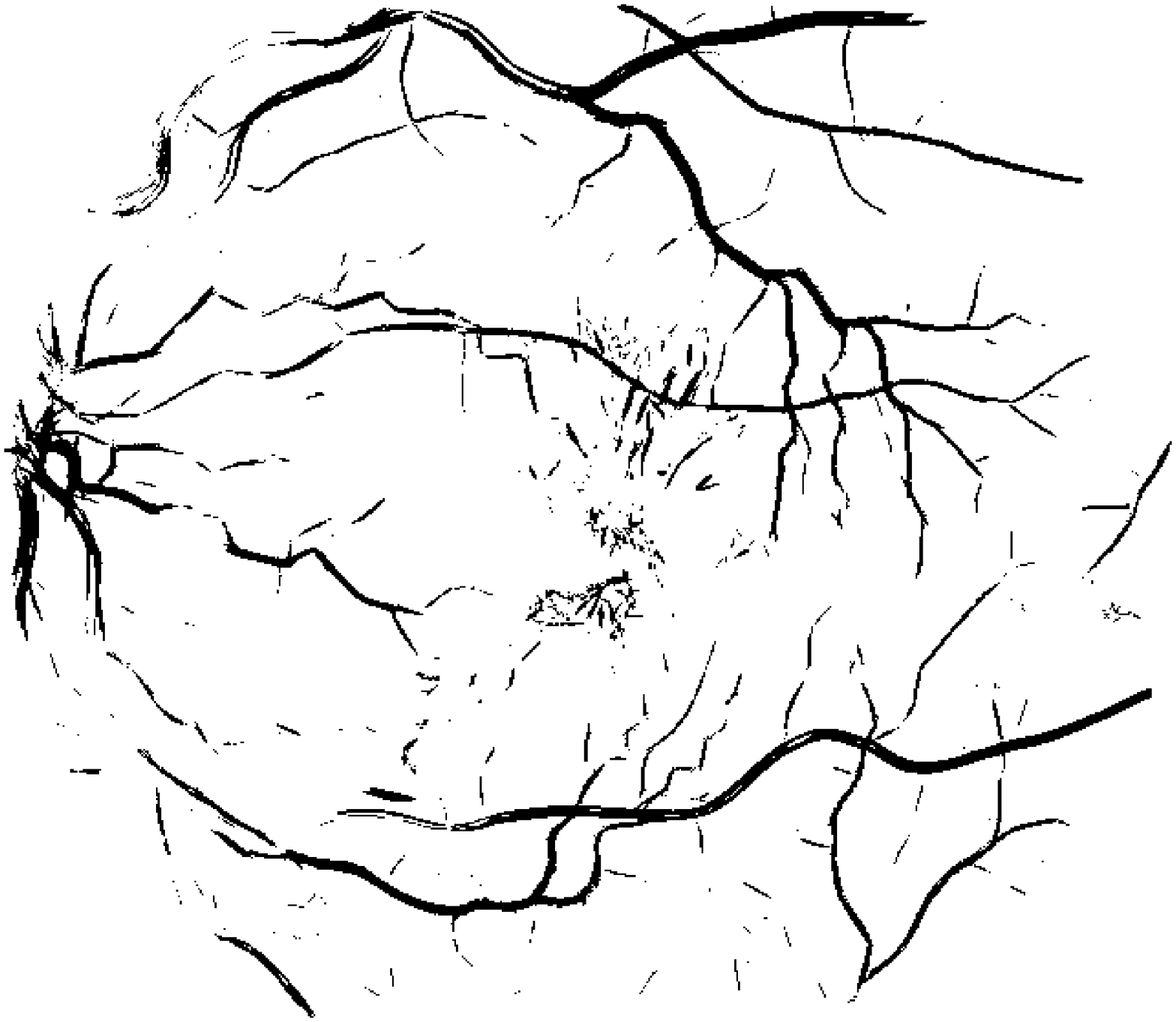}
    }
    \hfil
    \subfigure[First observer.]{\includegraphics[width=.23\textwidth]{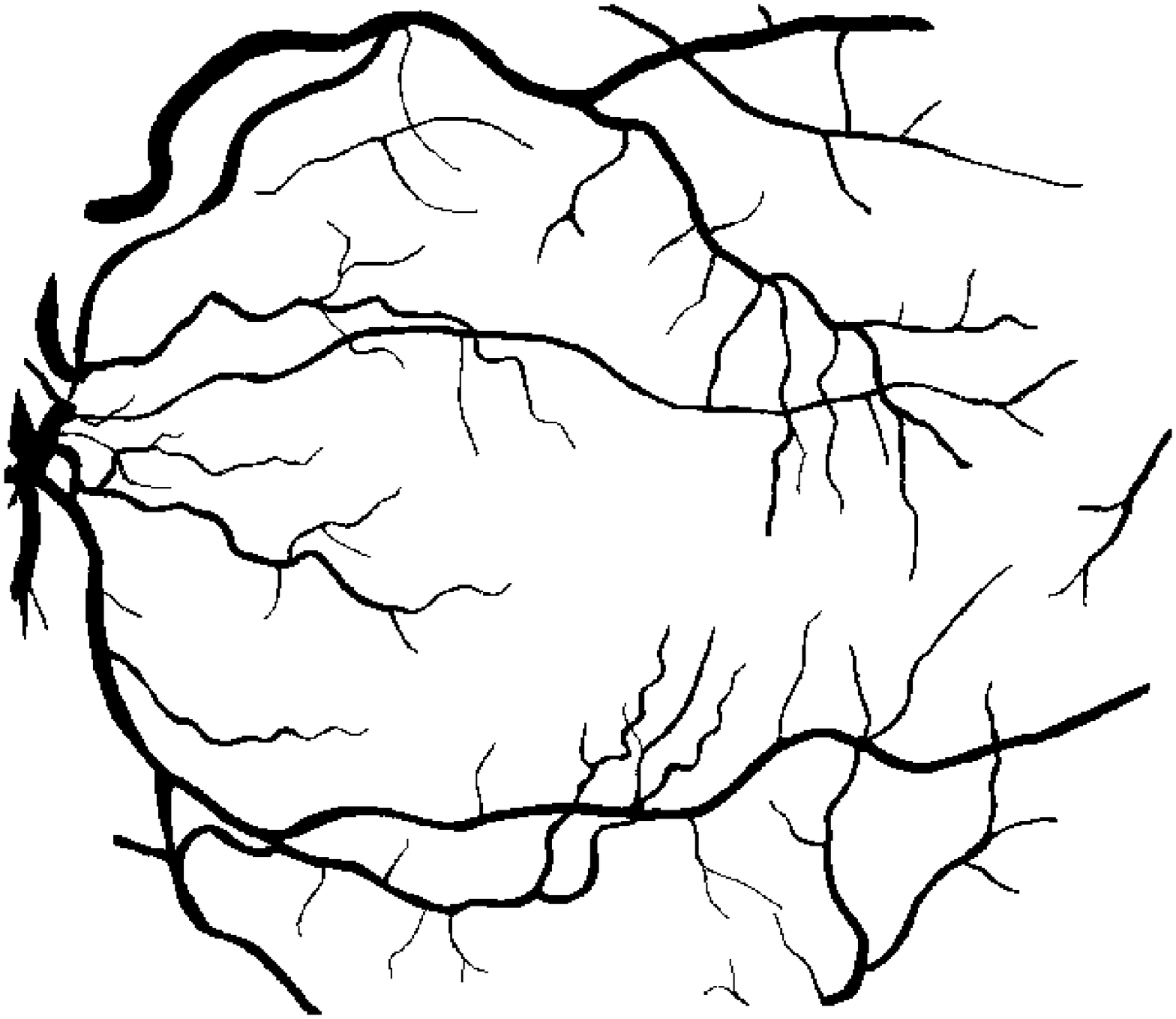}
    }
    \subfigure[Second observer.]{\includegraphics[width=.23\textwidth]{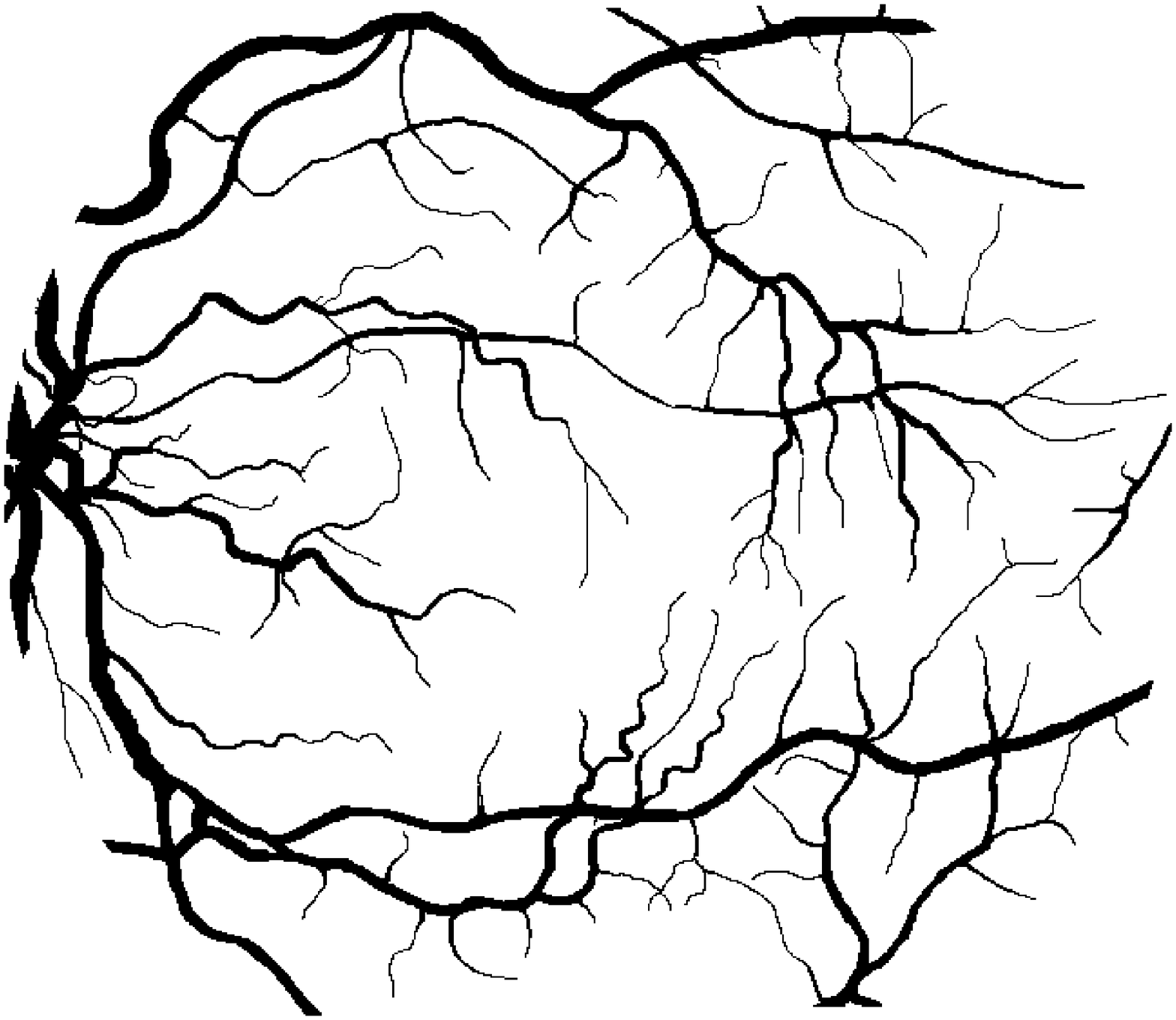}
    }
  }

  \centerline{\subfigure[Posterior probabilities.]{\includegraphics[width=.23\textwidth]{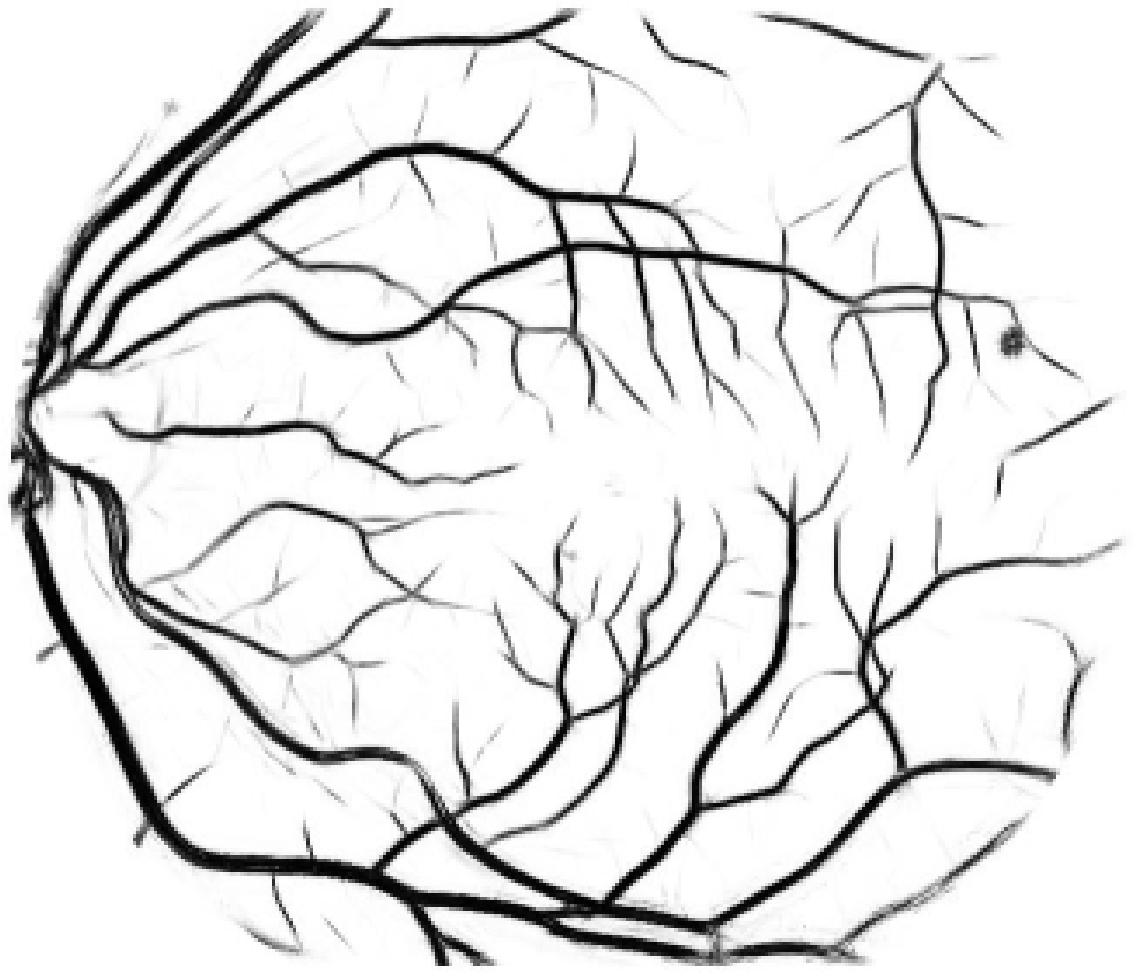}
    }
    \hfil
    \subfigure[Segmentation.]{\includegraphics[width=.23\textwidth]{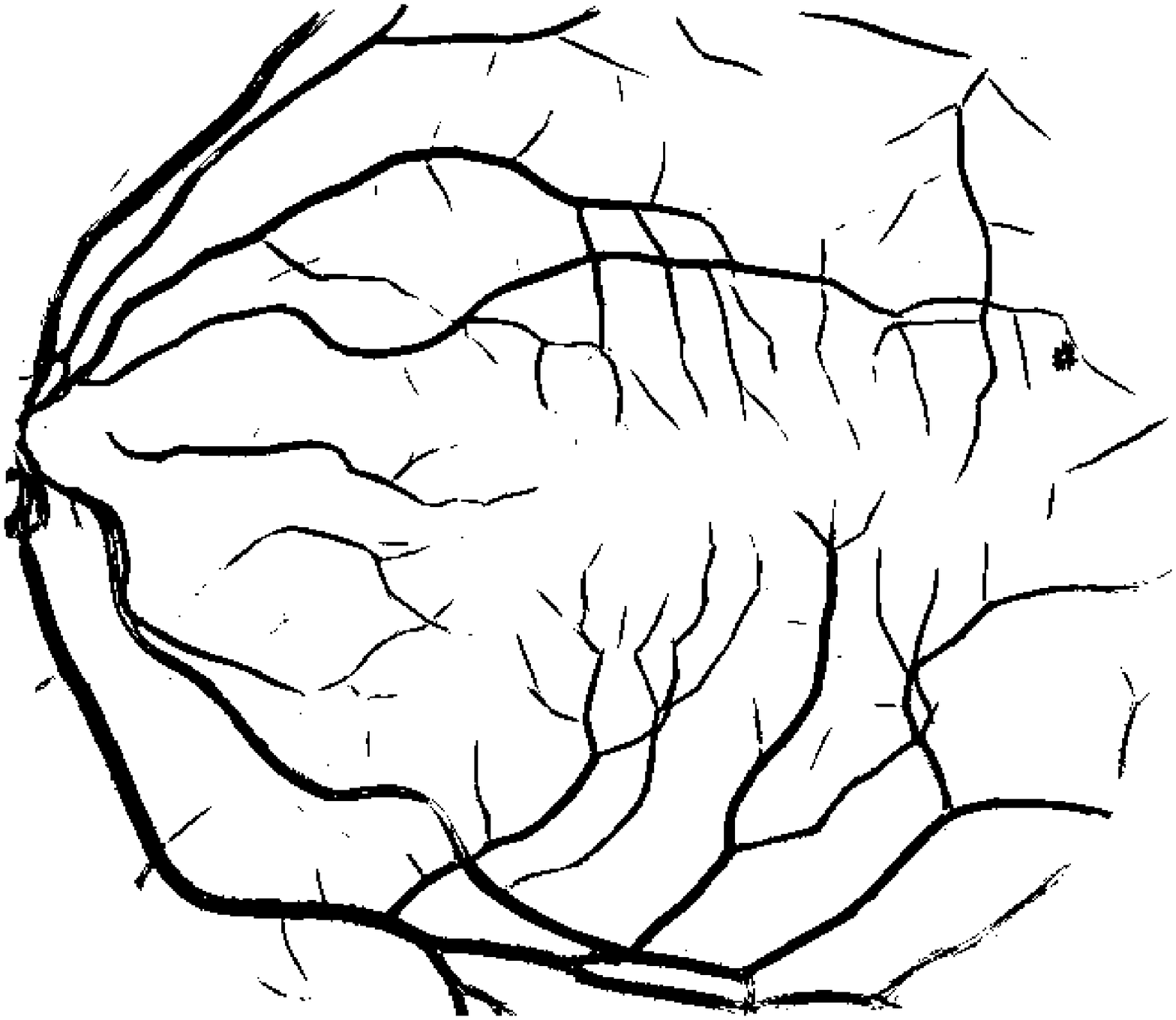}
    }
    \hfil
    \subfigure[First observer.]{\includegraphics[width=.23\textwidth]{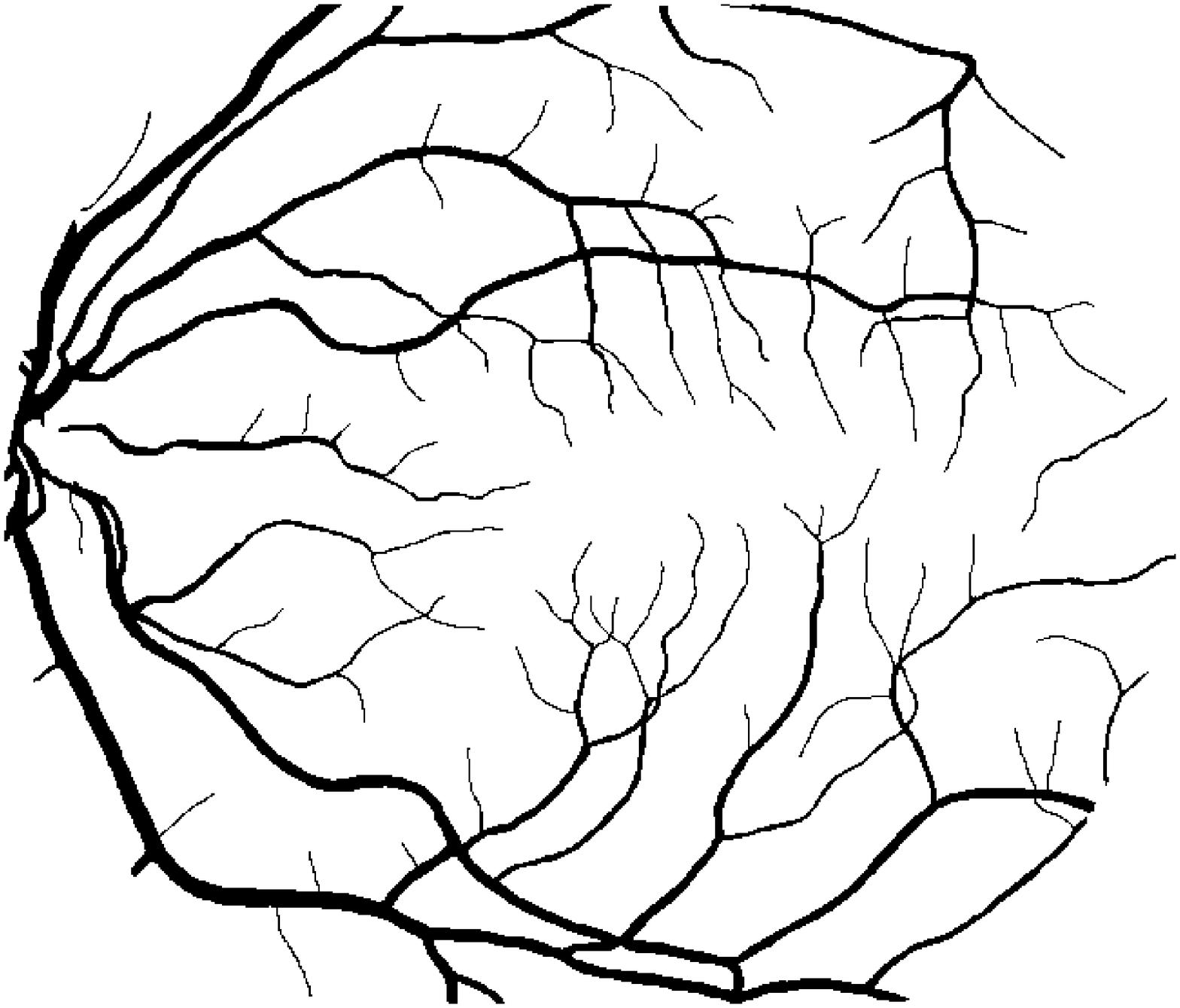}
    }
    \subfigure[Second observer.]{\includegraphics[width=.23\textwidth]{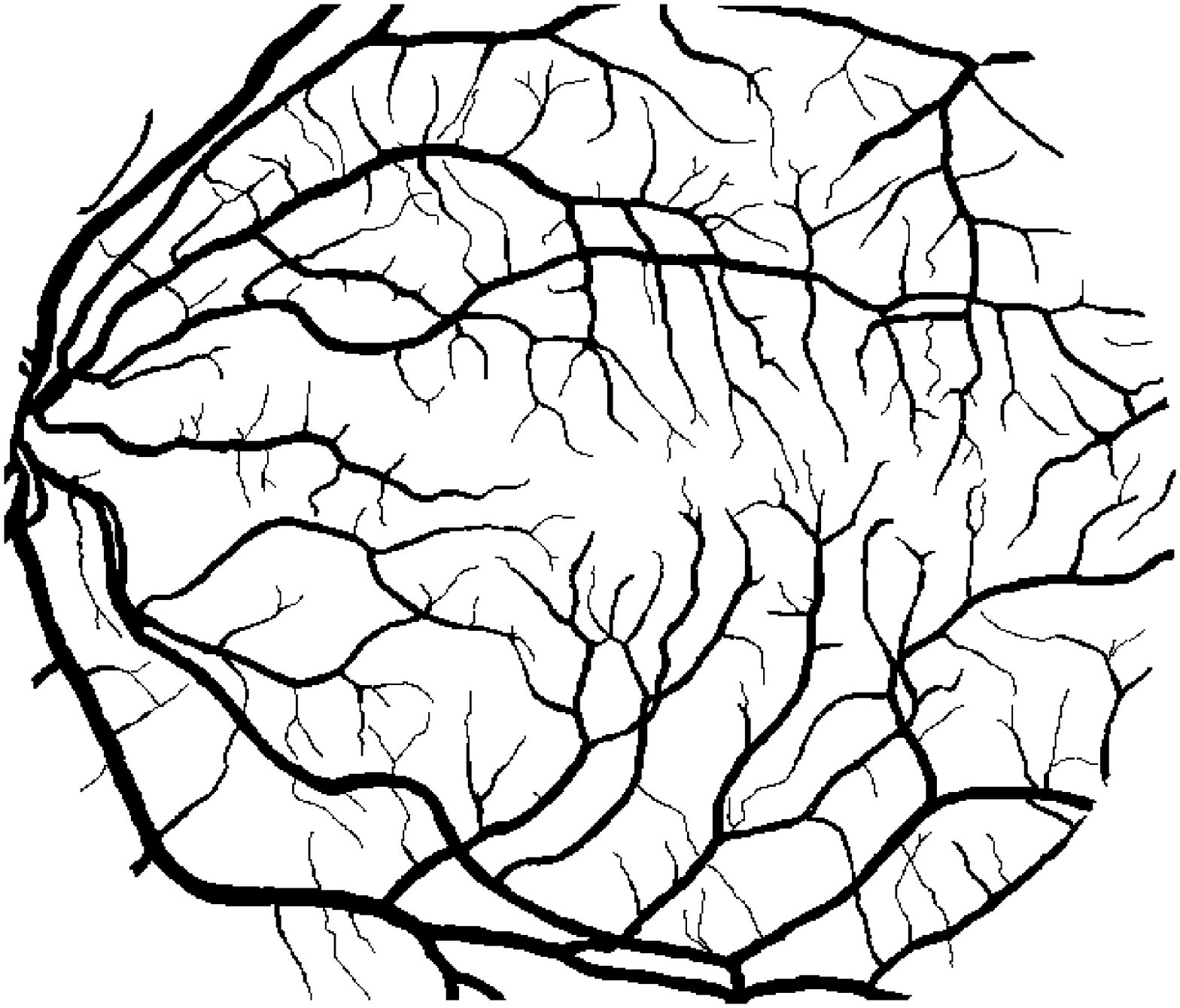}
    }
  }

  \caption{Results produced by the GMM classifier with $k = 20$ and
    manual segmentations for two images from the STARE database. The
    top row images originate from a pathological case, while the
    bottom ones originate from a normal case.}
  \label{result-stare}
\end{figure*}

Illustrative segmentation results for a pair of images from each
database (produced by the GMM classifier with $k = 20$), along with
the manual segmentations, are shown in Figs.~\ref{result-drive}
and~\ref{result-stare}.

\begin{figure}
\centering
\includegraphics[width=.45\textwidth]{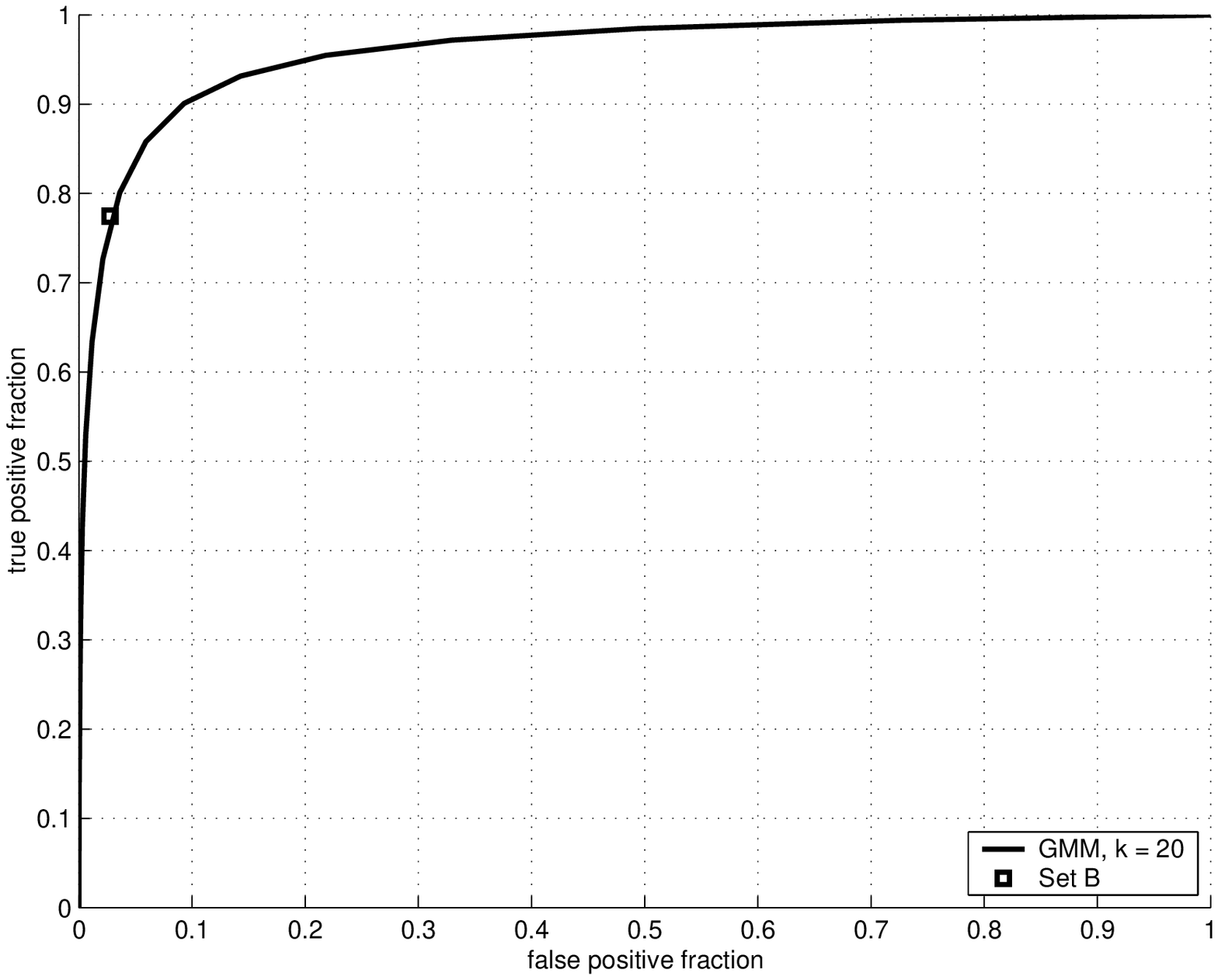}
\caption{ROC curve for classification on the DRIVE database using the
  GMM classifier with $k = 20$. The point marked
  as~\Squarepipe~corresponds to set B, the second set of manual
  segmentations. The method has $A_z = 0.9598$.}
\label{roc}
\end{figure}

\begin{figure}
\centering
\includegraphics[width=.45\textwidth]{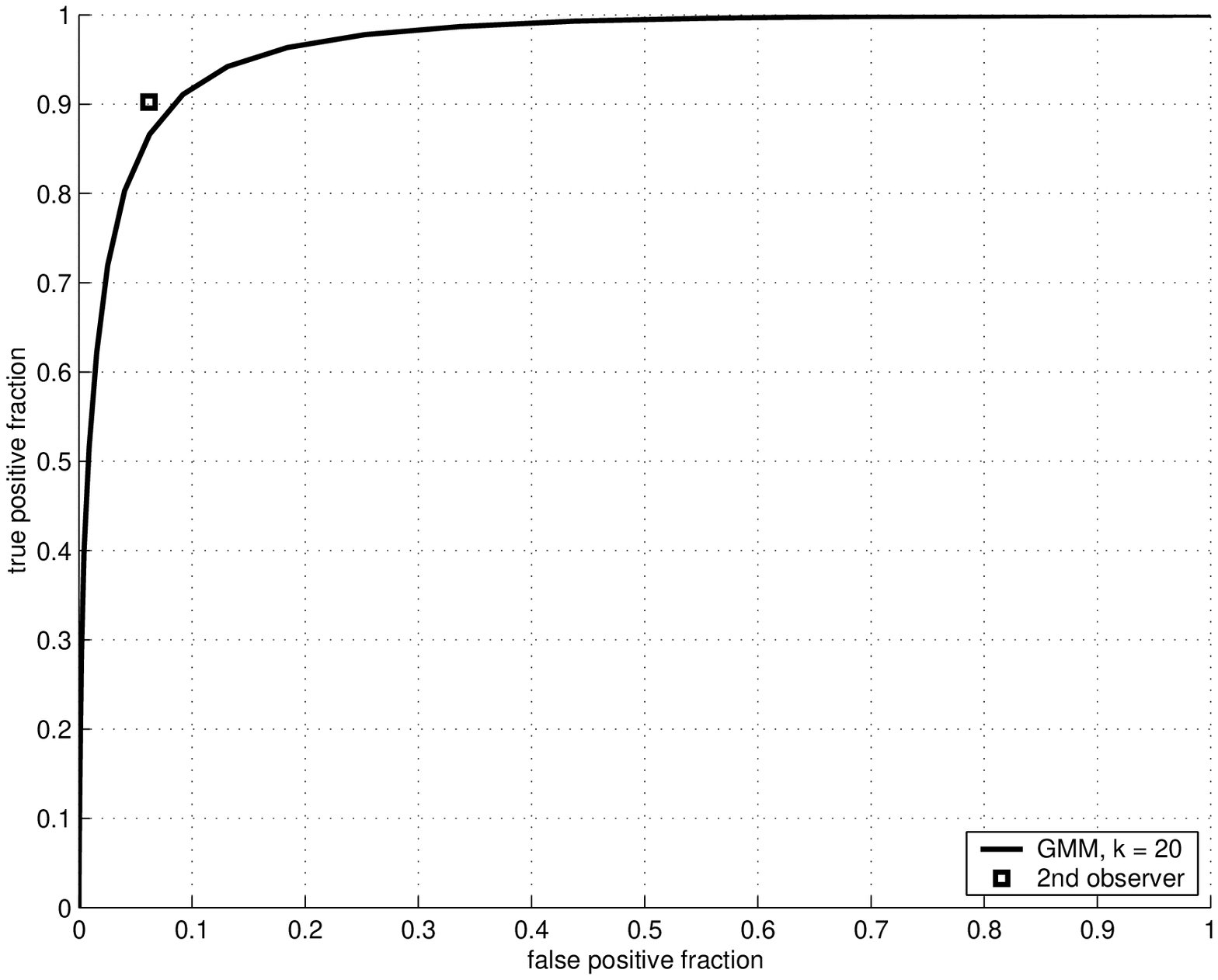}
\caption{ROC curve for classification on the STARE database using the
  GMM classifier with $k = 20$. The point marked
  as~\Squarepipe~corresponds to the second observer's manual
  segmentations.  The method has $A_z = 0.9651$.}
\label{roc-stare}
\end{figure}

For the DRIVE database, the manual segmentations from set A are used
as ground truth and the human observer performance is measured using
the manual segmentations from set B, which provide only one true/false
positive fraction pair, appearing as a point in the ROC graph
(Fig.~\ref{roc}). For the STARE database, the first observer's manual
segmentations are used as ground truth, and the second observer's
true/false positive fraction pair is plotted on the ROC graph
(Fig.~\ref{roc-stare}). The closer an ROC curve approaches the top
left corner, the better the performance of the method. A system that
agreed completely with the ground truth segmentations would yield an
area under the ROC curve $A_z = 1$. However, note that the second sets
of manual segmentations do not produce perfect true/false positive
fractions, for the manual segmentations evaluated disagree on some of
the pixels with the manual segmentations used as ground truth. Thus,
the variance between observers can be estimated, helping to set a goal
for the method's performance.

The areas under the ROC curves ($A_z$) are used as a single measure of
the performance of each method and are shown in Table~\ref{tabela} for
GMM classifiers of varying $k$ and for the LMSE classifier. For
comparison with the manual segmentations, we also measure the
accuracies (fraction of correctly classified pixels) of the automatic
and manual segmentations.  Note that the accuracy and $A_z$ values for
the GMM classifier increase with $k$.  The ROC curves for the DRIVE
and STARE databases produced using the GMM classifier with $k = 20$,
as well as performances for human observers, are shown in
Figs.~\ref{roc} and~\ref{roc-stare}.

We note that the EM training process for the GMMs is computationally
more expensive as $k$ increases, while the classification phase is
fast. On the other hand, LMSE is very fast for both training and
classification, but produces poorer results, as seen in
Table~\ref{tabela}.

\begin{table}
\renewcommand{\arraystretch}{1.3}
\caption{Results for different classification methods and human observer. $A_z$ indicates the area under the ROC curve, while the accuracy is the fraction of pixels correctly classified.}
\label{tabela}
\centering
\begin{tabular}{|c||c|c|c|c|}
\hline
& \multicolumn{4}{|c|}{Database}\\
\cline{2-5}
Classification Method & \multicolumn{2}{|c|}{DRIVE} & \multicolumn{2}{|c|}{STARE}\\
\cline{2-5}
 & $A_z$ & Accuracy & $A_z$ & Accuracy\\
\hline
LMSE & 0.9520 & 0.9280 & 0.9584 & 0.9362\\
GMM, $k = 1$ & 0.9250 & 0.9217 & 0.9394 & 0.9239\\
GMM, $k = 5$  & 0.9537 & 0.9431 &  0.9609 & 0.9430\\
GMM, $k = 10$  & 0.9570 & 0.9454 &  0.9627 & 0.9450\\
GMM, $k = 15$  & 0.9588 & 0.9459 & 0.9648 & 0.9470\\
GMM, $k = 20$  & 0.9598 &  0.9467 & 0.9651 & 0.9474\\
2nd. observer & & 0.9473 &  & 0.9349\\
\hline
\end{tabular}
\end{table}

\section{Discussion and conclusion}
\label{discussion}

The Morlet transform shows itself efficient in enhancing vessel
contrast, while filtering out noise. Information from Morlet
transforms at different scales, which allows the segmentation of
vessels of different diameters, are integrated through the use of the
statistical classifiers presented.  The LMSE classifier shows a
reasonable performance with a fast classification and training phase,
while the GMM classifier has a computationally demanding training
phase, but guarantees a fast classification phase and better
performance.

The classification framework demands the use of manual labelings, but
allows the methods to be trained for different types of images
(provided the corresponding manual segmentations are available),
possibly adjusted to specific camera or lighting conditions and are
otherwise automatic, i.e., adjustment of parameters or user
interaction is not necessary.  We are studying the use of training
sets composed of a small portion of the image to be segmented. Using
this approach, a semi-automated fundus segmentation software may be
developed, in which the operator only has to draw a small portion of
the vessels over the input image or simply click on several pixels
associated with the vessels. The remaining image would then be
segmented based on the partial training set. This approach is
interesting since it requires a small effort from the operator, which
is compensated by the fact that image peculiarities are directly
incorporated by the classifier.

It is curious to note that, on the STARE database, the accuracy of the
method is higher than that of the second observer
(Table~\ref{tabela}). The second observer's manual segmentations
contain much more of the thinnest vessels than the first observer
(lowering their accuracy), while the method, trained by the first
observer, is able to segment the vessels at a similar rate. However,
the ROC graph (Fig.~\ref{roc-stare}) still reflects the higher
precision of the second observer, due to some difficulties found by
the method, as discussed below.

It is possible to use only the skeleton of the segmentations for the
extraction of features from the vasculature. Depending on the
application, different evaluation methods become more appropriate
\cite{bowyer98}. For example, the evaluation of the skeleton would not
take into account the width of the vessels, but could measure other
qualities such as the presence of gaps and detection of branching
points.  Another interesting form of evaluation would be directly
through an application, such as in detection of neovascularization by
means of analysis and classification of the vessel
structure~\cite{cesarjelinek03}. A major difficulty in evaluating the
results is the establishment of a reliable ground
truth~\cite{fritzsche03}. Human observers are subjective and prone to
errors, resulting in large variability between observations. Thus, it
is desirable that multiple human-generated segmentations be combined
to establish a ground truth, which was not the case in the analysis
presented.

Though very good ROC results are presented, visual inspection shows
some typical difficulties of the method that must be solved by future
work. The major errors are in false detection of noise and other
artifacts. False detection occurs in some images for the border of the
optic disc, haemorrhages and other types of pathologies that present
strong contrast. Also, the method did not perform well for very large
variations in lighting throughout an image, but this occurred for only
one image out of the 40 tested from both databases. This could
possibly be solved by including intra-image normalization in the
pre-processing phase~\cite{cree-etal-2005b}. Another difficulty is the
inability to capture some of the thinnest vessels that are barely
perceived by the human observers.

Another drawback of our approach is that it only takes into account
information local to each pixel through image filters, ignoring useful
information from shapes and structures present in the image.  We
intend to work on methods addressing this drawback in the near future.
The results can be slightly improved through a post-processing of the
segmentations for removal of noise and inclusion of missing vessel
pixels as in~\cite{leandro03}. An intermediate result of our method is
the intensity image of posterior probabilities, which could possibly
benefit from a threshold probing as in~\cite{hoover00} or region
growing schemes.

Automated segmentation of non-mydriatic images provides the basis for
automated assessment by community health workers. Skeletonized images
of the vessel pattern of the ocular fundus can be analyzed
mathematically using nonlinear methods such as global
fractal~\cite{cesarjelinek03} and local fractal~\cite{mcquellin02}
analysis based on the wavelet transform thus providing a numeric
indicator of the extent of neovascularization. Our ongoing work aims
at applying the shape analysis and classification strategies described
in~\cite{cesarjelinek03} to the segmented vessels produced by method
described in this work.

\section*{Acknowledgments}

The authors thank J. J. Staal {\it et al.}~\cite{staal04} and A.
Hoover {\it et al.}~\cite{hoover00} for making their databases
publicly available and Dr. Alan Luckie and Chris McQuellin from the
Albury Eye Clinic for providing fluorescein images used during our
research.

\bibliographystyle{IEEEtran.bst}
\bibliography{IEEEabrv,ieee}

\end{document}